\documentclass[pdflatex,sn-mathphys]{sn-jnl}

\jyear{2021}%

\theoremstyle{thmstyleone}%
\theoremstyle{thmstyletwo}%

\theoremstyle{thmstylethree}%

\usepackage{graphicx}
\usepackage{subfigure}
\usepackage{amsmath}
\usepackage{longtable}
\usepackage{enumerate}
\usepackage{color}
\begin{document}

\title[Quantum image edge detection based on eight-direction Sobel operator for NEQR]{Quantum image edge detection based on eight-direction Sobel operator for NEQR }


\author*[1]{\fnm{Wenjie} \sur{Liu}}\email{wenjiel@163.com}

\author[2]{\fnm{Lu} \sur{Wang}}\email{Lu\_Wang\_MT@163.com}

\affil*[1]{\orgdiv{School of Computer Science}, \orgname{Nanjing University of Information Science  and Technology},
\orgaddress{ \city{Nanjing}, \postcode{210044}, \state{Jiangsu} \country{China}}}

\affil[2]{\orgdiv{School of Automation}, \orgname{Nanjing University of Information Science  and Technology}, \orgaddress{ \city{Nanjing}, \postcode{210044}, \state{Jiangsu}, \country{China}}}


\abstract{
 Quantum Sobel edge detection (QSED) is a kind of algorithm for image edge detection using quantum mechanism, which can solve the real-time problem encountered by classical algorithms. However, the existing QSED algorithms only consider two- or four-direction  Sobel operator, which leads to a certain loss of edge detail information in some high-definition images. In this paper, a novel QSED algorithm based on eight-direction Sobel operator is proposed, which not only reduces the loss of edge information, but also simultaneously calculates eight directions' gradient values of all pixel in a quantum image. In addition, the concrete quantum circuits, which 
consist of gradient calculation, non-maximum  suppression,  double  threshold  detection  and  edge tracking units, are designed in details. For a  ${2^n} \!\times\! {2^n}$ image with $q$ gray-scale, the  complexity  of our  algorithm can be reduced to O(${n^2} + {q^2}$), which is lower than   other existing classical or quantum algorithms.    And the simulation experiment  demonstrates that our  algorithm can detect more edge information, especially  diagonal
edges, than the two- and four-direction QSED algorithms.}




\maketitle

\section{Introduction}\label{Section 1}

In recent years, quantum image processing (QIP) received widespread attention and deep research from researchers  as an emerging sub-discipline of  quantum computing 
and image processing \cite{Yan2017,Cai2018}. Due to the parallelism and entanglement properties of quantum computing, the computational speed can be improved in different degrees than classical computing in some problems. At present, the demand for high-quality images is increasing,  which  results in a sharp increase in computation time on classical computers. Therefore, the  real-time problem of digital image processing encounters great challenges. Quantum image processing can utilize the advantages of quantum computing to improve the processing speed, which makes it necessary to develop image processing  on quantum computers.

Quantum image processing is usually divided into three stages: quantum image representation, quantum image processing algorithm and measuring quantum image information. Quantum image representation is a model that represents digital images as quantum images. At present, there are many quantum image representation models and can be approximately divided into two categories \cite{Yan2016A,Yan2022}. One is to encode the gray-scale values of the quantum image into the probability amplitude of the qubits, which can encode images using fewer qubits, such as qubit lattice representation \cite{Venegas-Andraca2003}, real ket representation \cite{Latorre2005}, entangled images representation \cite{Venegas-Andraca2010},   flexible representation of quantum image (FRQI) \cite{Le2011},   multi-channel RGB images representation of quantum images (MCQI) \cite{Sun2013}, normal arbitrary quantum superposition state (NAQSS) \cite{Li2014}, quantum probability image encoding representation (QPIE) \cite{Yao2018}. When an image is retrieved, a large number of measurements are required to get an approximation of the probability magnitude, which makes it difficult to retrieve images.   The other method, such as  novel enhanced quantum representation (NEQR) \cite{Zhang2013}, novel quantum representation of color digital images (NCQI) \cite{Sang2017} and novel quantum image representation based on HSI color space model (QIRHSI) \cite{Chen2022}, solves this problem well and it is to encode the gray-scale values by using a separate qubit sequence.  When an image is retrieved, the gray-scale value of each pixel can be accurately recovered with a few measurements.  Therefore, the NEQR model is widely used due to its simplicity of operation.  As different quantum image representation models 
are proposed, a large number of quantum image processing algorithms emerge, such as geometrical transformation of quantum image \cite{Le2010},  feature extraction of quantum image \cite{Hancock2015}, quantum image watermarking \cite{Song2014},  quantum image bilinear interpolation \cite{Yan2021}, quantum image segmentation \cite{Xia2019,Yuan2020}, quantum image steganography \cite{Zhao2021}, quantum image edge detection \cite{Zhang2015,Fan2019}, etc. 
  
Image edge detection is a fundamental problem in image processing, which can retain the basic framework in the image,  remove irrelevant information and reduce the amount of data. Currently, the digital image edge detection algorithms have been widely explored, but the research on quantum counterparts is still in its infancy. Many researchers use different operators, such as  Sobel \cite{Zhang2015}, Prewitt \cite{Zhou2019P}, LoG \cite{Li2020}, etc. for edge detection of quantum images,  among which the Sobel operator is the first choice. In 2015, Zhang et al. \cite{Zhang2015} firstly proposed a quantum Sobel edge detection (QSED) algorithm for FRQI image.  This algorithm is a quantization of the classical Sobel edge detection by using quantum circuit, and there is an exponential acceleration relative to the classical method, which improves the real-time performance, but it can not  accurately measure the color information of the image. In 2019, Fan et al. \cite{Fan2019} proposed  a two-direction QSED algorithm for  NEQR  image.  However, only the edges in the horizontal and vertical directions were considered in their algorithm, which exists large limitations. In order to improve the accuracy of edge detection, Zhou et al. \cite{Zhou2019} proposed a four-direction QSED algorithm for generalized quantum image representation (GQIR) image,   but its circuit complexity is higher than other algorithms. For NEQR image, Chetia et al. \cite{Chetia2019} also proposed a four-direction QSED algorithm, but  the edge detection effect of this algorithm is relatively poor. In order to detect more edge information and reduce circuit complexity,   they \cite{Chetia2021} furture proposed an improved version  in 2021. 
As far as we know,  the
existing  QSED algorithms only consider either two- or four-direction Sobel operator, and  the edge information detected is insufficient in some scenarios.
Therefore, we have done further research on  QSED,  and the main works  are summarized as follows:
\begin{itemize}
    \item A  NEQR  image edge detection algorithm based on eight-direction Sobel operator  is proposed.
    
    \item Several specific quantum circuits are designed, which can  simultaneously calculate  eight directions' gradient values of all pixels,   and  classify the pixels accurately according to the obtained gradient values.
    
    \item  We verify the superiority and feasibility of our proposed algorithm by analyzing the circuit complexity and performing  simulation experiments, respectively.
\end{itemize}

This paper is organized as follows. Sec. \ref{Section 2} introduces the principle of the NEQR representation model and the classical edge detection of eight-direction Sobel operator. In Sec. \ref{Section 3}, some basic quantum operation modules are introduced. Then, a series of quantum circuits  of edge detection  are designed and the relevant quantum states equations are given. Sec. \ref{Section 4} analyzes the computational complexity of our algorithm and experimental results, and the conclusion is drawn in Sec. \ref{Section 5}.

\section{Related work}\label{Section 2}
\subsection{NEQR}

A pixel in a digital image contains the position  and color information. The NEQR  uses two entangled qubit sequences to store the grayscale information and position information of the image, and stores the entire image in a superposition of the two-qubit sequences. For grayscale images of size ${2^n} \times {2^n}$, the grayscale range is $[0,{2^q} - 1]$ and requires a qubit sequence of $q$ length to store the grayscale of pixels. Moreover,   two qubit sequences of $n$ length are needed  to store the  position information of each pixel in the image. The entire representation is the tensor product of these three entangled qubits sequences, so that all pixels can be stored and processed simultaneously. Then the NEQR model of a quantum image can be written in the form of the quantum superposition state shown in Eq. (1) \cite{Zhang2013}.
\begin{equation}
   \lvert {\rm{I}} \rangle  = \frac{1}{{{2^n}}}\sum\limits_{Y = 0}^{{2^n} - 1} {\sum\limits_{X = 0}^{{2^n} - 1} {\lvert {{C_{YX}}} \rangle  \otimes \lvert Y \rangle \lvert X \rangle } }  = \frac{1}{{{2^n}}}\sum\limits_{YX = 0}^{{2^{2n}} - 1} {\mathop  \otimes \limits_{k = 0}^{q - 1}\lvert {C^K_{YX}} \rangle \mathop  \otimes \limits_{}^{} \lvert {YX} \rangle  } 
\end{equation}
where  $\lvert {{C_{YX}}} \rangle  = \lvert {C_{YX}^{q - 1},C_{YX}^{q - 2},{ \cdots ^{}}C_{YX}^{1}C_{YX}^{0}} \rangle$ represents the quantum image gray-scale values, $C_{YX}^k \in \left\{ {0,1} \right\}$, $k = q - 1,q - 2, \cdots ,0$. $  \lvert {YX} \rangle  = \lvert Y \rangle \lvert X\rangle  = \lvert {{Y_{n - 1}},{Y_{n - 2}}, \cdots {Y_0}}\rangle \lvert {{X_{n - 1}},{X_{n - 2}}, \cdots {X_0}} \rangle $ represents the 
position of the pixel in a quantum image, ${Y_t},{X_t} \in \left\{ {0,1} \right\}$.

\begin{figure}
 \centering
    \includegraphics[width=2cm]{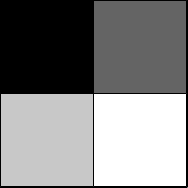}
    \caption{An example of a 2×2 image}
    \label{figure l}
\end{figure}

Fig. \ref{figure l} shows an example of a grayscale image of size 2×2, and the corresponding NEQR expression of which is given as follows:

\[\begin{array}{l}
\lvert I \rangle  = \frac{1}{2}\left( {\lvert 0 \rangle \lvert{00} \rangle  + \lvert {100} \rangle \lvert {01} \rangle  + \lvert {200} \rangle \lvert {10} \rangle  + \lvert{255} \rangle \lvert {11} \rangle } \right)
\end{array}\]
$$\begin{array}{l}
\begin{array}{*{20}{c}}
{}& = 
\end{array}\frac{1}{2}\left( \begin{array}{l}
\lvert {00000000} \rangle \lvert{00} \rangle  + \lvert {01100100} \rangle \lvert {01} \rangle \\
 + \lvert {11001000}\rangle \lvert{10} \rangle  + \lvert {11111111} \rangle \lvert {11} \rangle 
\end{array} \right)
\end{array}
\eqno{(2)}$$

\subsection{Classical Sobel edge detection}

\begin{figure}
    \centering
    \includegraphics[width=6cm]{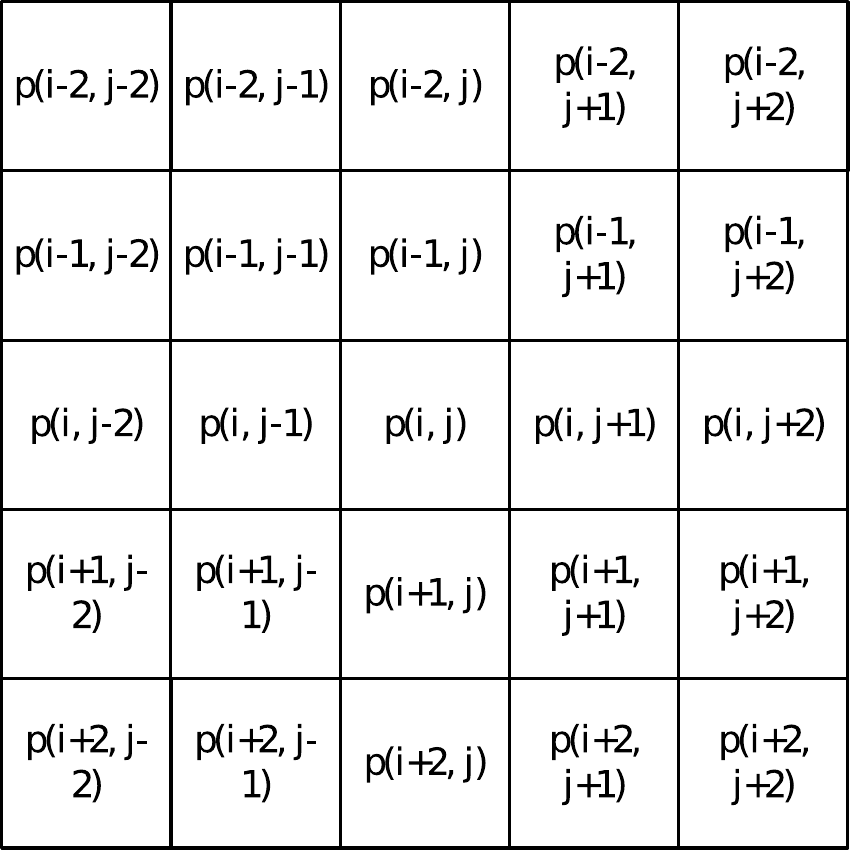}
    \caption{The pixel neighborhood window}
    \label{fig 2}
\end{figure}

Image's edges are caused by discontinuous color intensities, and they are the pixels where the color intensity of the image changes the fastest.
 Based on this principle, many operators for edge detection appear. Among them, the Sobel operator is the most widely used. The Sobel operator consists of a set of masks of size $3\times3$ to calculate the gradient of pixel color intensity in the image. The underlying Sobel operator has two  directions and can calculate the horizontal and vertical gradients of the pixels. This detects the horizontal and vertical edges in the image, but the calculated image edges are rectangular. So the detected edges can be further improved. Therefore, to obtain better edge detection effects, the underlying Sobel operator can be rotated to obtain the Sobel operators in four directions of ${0^ \circ }$, ${45^ \circ }$, ${90^ \circ }$and ${135^ \circ }$. The Sobel operator with $3\times3$ in four directions can detect the edges of the image more accurately. But the detected edge continuity is not   enough. To detect the edge pixels in the image more accurately, the Sobel operator can be improved to  eight directions: ${0^ \circ }$, ${22.5^ \circ }$, ${45^ \circ }$, ${67.5^ \circ }$ , ${90^ \circ }$, ${112.5^ \circ }$, ${135^ \circ }$, ${157.5^ \circ }$ \cite{Zheng2013}. In addition, coupled with non-maximum suppression and edge tracking, the edge information will be  more precise and detailed \cite{Chetia2021}. Fig. \ref{fig 2} shows a $5\times5$  pixel neighborhood window. The pixels' gradient values in eight directions can be calculated by the following equations:

\begin{small}

\[{G^0} = \left[ {\begin{array}{*{20}{c}}
0&0&0&0&0\\
{ - 1}&{ - 2}&{ - 4}&{ - 2}&{ - 1}\\
0&0&0&0&0\\
1&2&4&2&1\\
0&0&0&0&0
\end{array}} \right] * p, {G^{22.5}} = \left[ {\begin{array}{*{20}{c}}
0&0&0&0&0\\
0&{ - 2}&{ - 4}&{ - 2}&0\\
{ - 1}&{ - 4}&0&4&1\\
0&2&4&2&0\\
0&0&0&0&0
\end{array}} \right] * p
\]

\[
{G^{45}} = \left[ {\begin{array}{*{20}{c}}
0&0&0&{ - 1}&0\\
0&{ - 2}&{ - 4}&0&1\\
0&{ - 4}&0&4&0\\
{ - 1}&0&4&2&0\\
0&1&0&0&0
\end{array}} \right] * p,
{G^{67.5}} = \left[ {\begin{array}{*{20}{c}}
0&0&{ - 1}&0&0\\
0&{ - 2}&{ - 4}&2&0\\
0&{ - 4}&0&4&0\\
0&{ - 2}&4&2&0\\
0&0&1&0&0
\end{array}} \right] * p
\]
$$
{G^{90}} = \left[ {\begin{array}{*{20}{c}}
0&{ - 1}&0&1&0\\
0&{ - 2}&0&2&0\\
0&{ - 4}&0&4&0\\
0&{ - 2}&0&2&0\\
0&{ - 1}&0&1&0
\end{array}} \right] * p,
{G^{112.5}} = \left[ {\begin{array}{*{20}{c}}
0&0&1&0&0\\
0&{ - 2}&4&2&0\\
0&{ - 4}&0&4&0\\
0&{ - 2}&{ - 4}&2&0\\
0&0&{ - 1}&0&0
\end{array}} \right] * p
\eqno{(3)}$$
\[{G^{135}} = \left[ {\begin{array}{*{20}{c}}
0&1&0&0&0\\
{ - 1}&0&4&2&0\\
0&{ - 4}&0&4&0\\
0&{ - 2}&{ - 4}&0&1\\
0&0&0&{ - 1}&0
\end{array}} \right] * p, {G^{157.5}} = \left[ {\begin{array}{*{20}{c}}
0&0&0&0&0\\
0&2&4&2&0\\
{ - 1}&{ - 4}&0&4&1\\
0&{ - 2}&{ - 4}&{ - 2}&0\\
0&0&0&0&0
\end{array}} \right] * p\]
\end{small}

Among these, ${G^0}$, ${G^{22.5}}$, ${G^{45}}$, ${G^{67.5}}$, ${G^{90}}$, ${G^{112.5}}$, ${G^{135}}$ and ${G^{157.5}}$ represent the image gradient values detected by the eight directional edges of ${0^ \circ }$, ${22.5^ \circ }$, ${45^ \circ }$, ${67.5^ \circ }$ , ${90^ \circ }$, ${112.5^ \circ }$, ${135^ \circ }$, ${157.5^ \circ }$, respectively. The $p$ represents a pixel neighborhood window. Specific calculations are as follows:
\begin{small}
\begin{small}
\[\begin{array}{l}
{G^0} \!= \!p(Y\!\! -\!\! 2,X\!\! +\!\! 1)\!\! + \!\!2p(Y\!\! -\!\! 1,X\!\! + \!\!1) \!\!+\!\! 4p(Y,X \!\!+ \!\!1)\!\! +\!\! 2p(Y\!\! + \!\!1,X\!\! + \!\!1) \!\!+ p(Y\!\! +\!\! 2,X\!\! + \!\!1)\\
\begin{array}{*{20}{c}}
{}&{}
\end{array} - \!\!p(Y\!\! - \!\!2,X \!\!-\!\! 1)\!\! -\!\! 2p(Y\!\! -\!\! 1,X \!\!- \!\!1)\!\! - \!\!4p(Y,X\!\! - \!\!1)\!\! - 2p(Y\!\! +\!\! 1,X\!\! - \!\!1) \!\!- p(Y\!\! +\!\! 2,X\!\! -\!\! 1)
\end{array}\]
\[\begin{array}{l}
{G^{22.5}} \!= \!p(Y\! +\! 2\!,X) \!+ \!2p(Y \!+ \!1,X\! +\! 1)\! +\! 2p(Y \!- \!1,X \!+\! 1)\! + \!4p(Y\!,X\! +\! 1) \!+\! 4p(Y\! +\! 1,X)\\
\begin{array}{*{20}{c}}
{}&{}
\end{array} - \!p(Y\! - \!2,X)\! -\! 2p(Y \!+ \!1,X\! - \!1)\! -\! 2p(Y\! - \!1,X \!- \!1)\! - \!4p(Y,X\! -\! 1) \!-\! 4p(Y\! - \!1,X)
\end{array}\]
\[\begin{array}{l}
{G^{45}}\! =\! p(Y\! +\! 2,X \!-\! 1)\! +\! p(Y\! - \!1,X\! +\! 2) \!+\! 2p(Y\! +\! 1,X \!+\! 1) \!+ \!4p(Y\! +\! 1,X)\! + \!4p(Y,X \!+ \!1)\\
\begin{array}{*{20}{c}}
{}&{}
\end{array} - \!p(Y \!+ \!1,X\! -\! 2) \!- \!p(Y\! -\! 2,X\! +\! 1) \!- \!2p(Y\! -\! 1,X \!-\! 1) \!-\! 4p(Y\! -\! 1,X) \!- \!4p(Y,X \!-\! 1)
\end{array}\]
\[\begin{array}{l}
{G^{67.5}} \!= \!p(Y,X\!+\!2)\!+\!2p(Y\!+\!1,X\!+\!1)\!+\!2p(Y\!+\!1,X\!-\!1)\! +\!4p(Y\!+\!1,X)\!+\!4p(Y,X\!+\!1)\\
\begin{array}{*{20}{c}}
{}&{}
\end{array} -\! p(Y,X\! -\! 2) \!-\! 2p(Y\! - \!1,X\! +\! 1)\! - \!2p(Y \!-\! 1,X\! - \!1) \!-\! 4p(Y \!- \!1,X)\! - \!4p(Y,X\! -\! 1)
\end{array}\]
\[\begin{array}{l}
{G^{90}} \!= \!p(Y\! + \!1\!,\!X \!- \!2)\! +\! p(Y\! + \!1\!,\!X \!+\! 2)\! +\! 2p(Y\! +\! 1\!,\!X \!-\! 1)\! + \!2p(Y \!+ \!1\!,\!X\! +\! 1)\! +\! 4p(Y \!+ \!1\!,\!X)\\
\begin{array}{*{20}{c}}
{}&{}
\end{array} -\! p(Y \!- \!1,X\! -\! 2) \!-\! p(Y\! -\! 1\!,X\! +\! 2) \!- \!2p(Y\! -\! 1\!,\!X \!- \!1)\! -\! 2p(Y\! -\! 1\!,\!X\! + \!1) \!-\! 4p(Y \!
-\! 1\!,\!X)
\end{array}\]
\[\begin{array}{l}
{G^{112.5}} \!= \!p(Y,X\! -\! 2) \!+ \!2p(Y\! +\! 1,X \!- \!1) \!+\! 2p(Y \!+\! 1,X\! +\! 1) \!+\! 4p(Y\! + \!1,X) \!+\! 4p(Y,X \!- \!1)\\
\begin{array}{*{20}{c}}
{}&{}
\end{array} - \!p(Y,X \!+\! 2) \!- \!2p(Y\! -\! 1,X\! + \!1)\! -\! 2p(Y\! - \!1,X \!- \!1)\! -\! 4p(Y \!-\! 1,X)\! -\! 4p(Y,X \!+ \!1)
\end{array}\]
\[\begin{array}{l}
{G^{135}} \!= \!p(Y \!- \!1\!,\!X \!- \!2)\! +\! p(Y\! +\! 1\!,\!X\! +\! 1)\! +\! 2p(Y\! + \!1\!,\!X\! -\! 1) \!+\! 4p(Y\! + \!1\!,\!X)\! + \!4p(Y\!,\!X\! -\! 1)\\
\begin{array}{*{20}{c}}
{}&{}
\end{array} -\! p(Y \!- \!2\!,\!X \!- \!1)\! \!-\! p(Y\! +\! 1,X\! +\! 2) \!- \!2p(Y\! -\! 1\!,\!X \!+\! 1) \!-\! 4p(Y\! - \!1\!,\!X) \!- \!4p(Y\!,\!X \!+ \!1)
\end{array}\]
$$\begin{array}{l}
{G^{157.5}}\! =\! p(Y \!+\! 2,X) \!+ \!2p(Y\! +\! 1,X \!-\! 1) \!+\! 2p(Y\! -\! 1,X \!-\! 1)\! +\! 4p(Y\! +\! 1,X) \!+ \!4p(Y,X\! - \!1)\\
\begin{array}{*{20}{c}}
{}&{}
\end{array} -\! p(Y \!- \!2,X)\! -\! 2p(Y\! +\! 1,X \!+\! 1) \!-\! 2p(Y \!-\! 1,X\! +\! 1) \!- \!4p(Y\! -\! 1,X) \!- \!4p(Y,X\! + \!1)
\end{array}
\eqno{(4)}$$
\end{small}
\end{small}

The gradient  of each pixel is the maximum of the absolute value of the gradient value in eight directions. It can be written as follows:
  $$ G = \max \left\{ {\lvert {{G^0}} \rvert,\lvert {{G^{22.5}}} \rvert,\lvert {{G^{45}}} \rvert,\lvert {{G^{67.5}}} \rvert,\lvert {{G^{90}}} \rvert,\lvert {{G^{112.5}}} \rvert,\lvert {{G^{135}}} \rvert,\lvert {{G^{157.5}}} \rvert} \right\} \eqno{(5)}$$

\begin{figure}
    \centering
   \subfigure[]{\includegraphics[width=2.5cm]{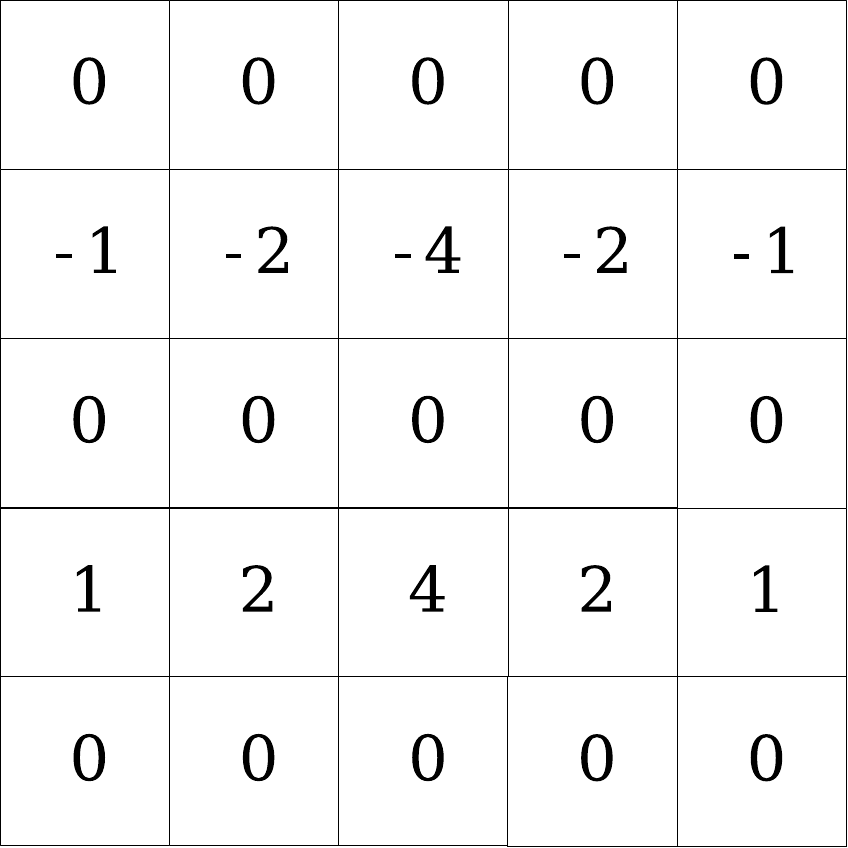}}
   \subfigure[]{\includegraphics[width=2.5cm]{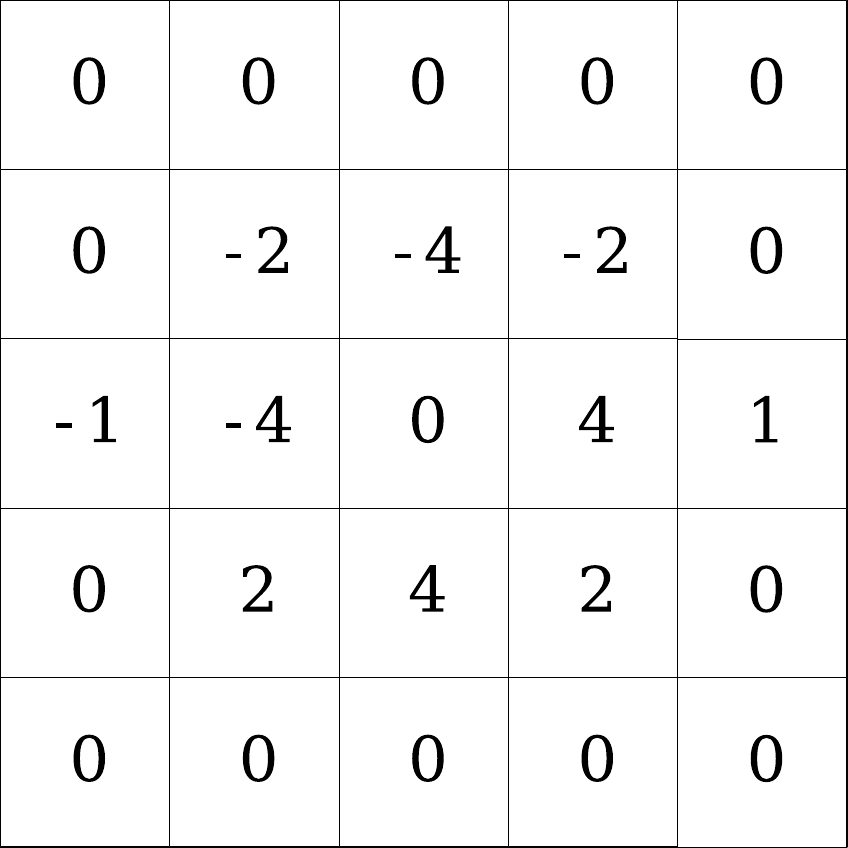}}
   \subfigure[]{\includegraphics[width=2.5cm]{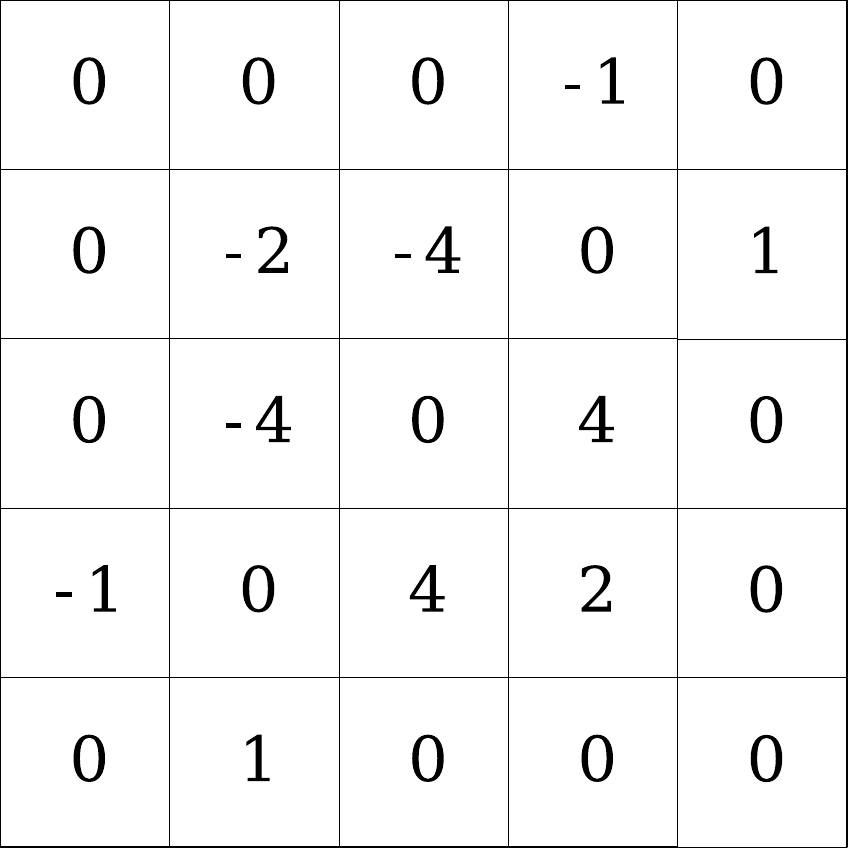}}
   \subfigure[]{\includegraphics[width=2.5cm]{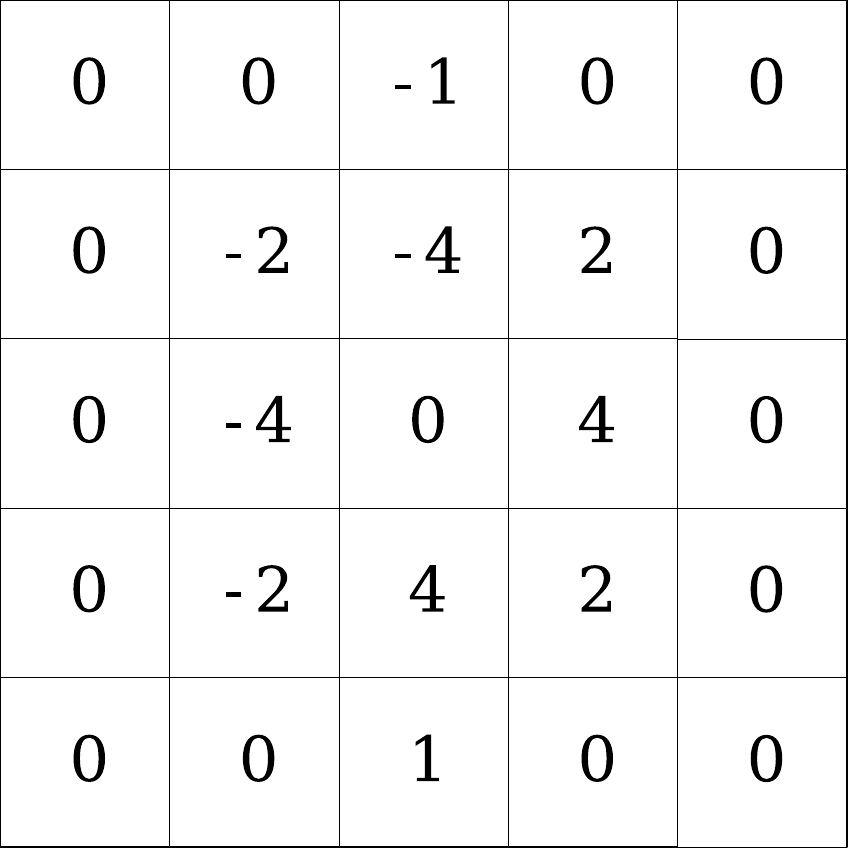}}
   \\
   \subfigure[]{\includegraphics[width=2.5cm]{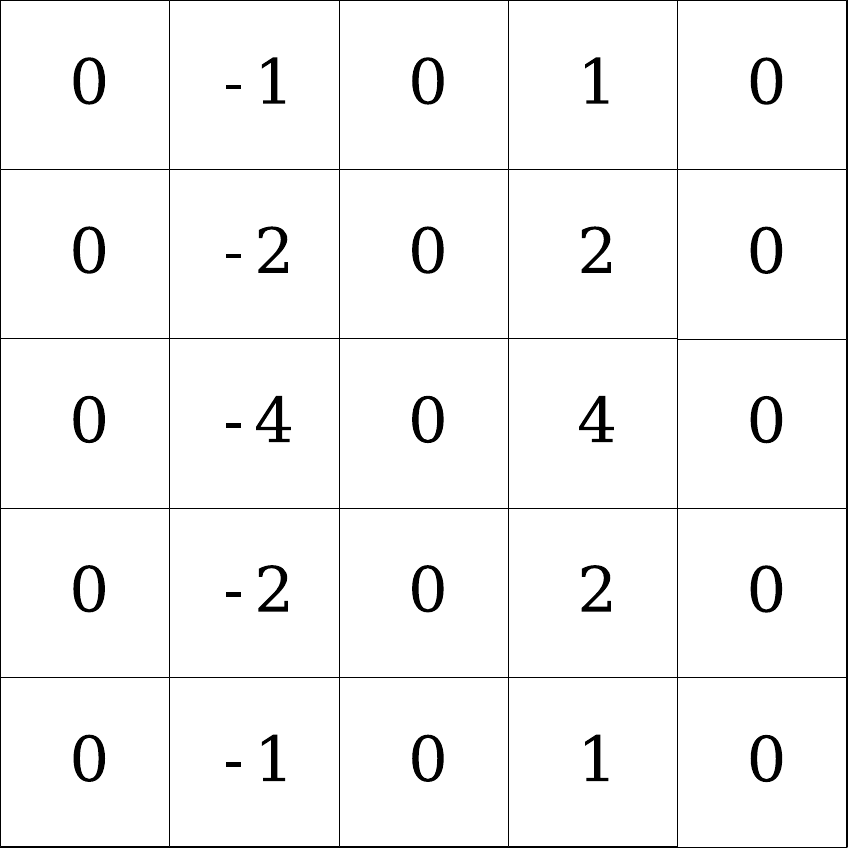}}
   \subfigure[]{\includegraphics[width=2.5cm]{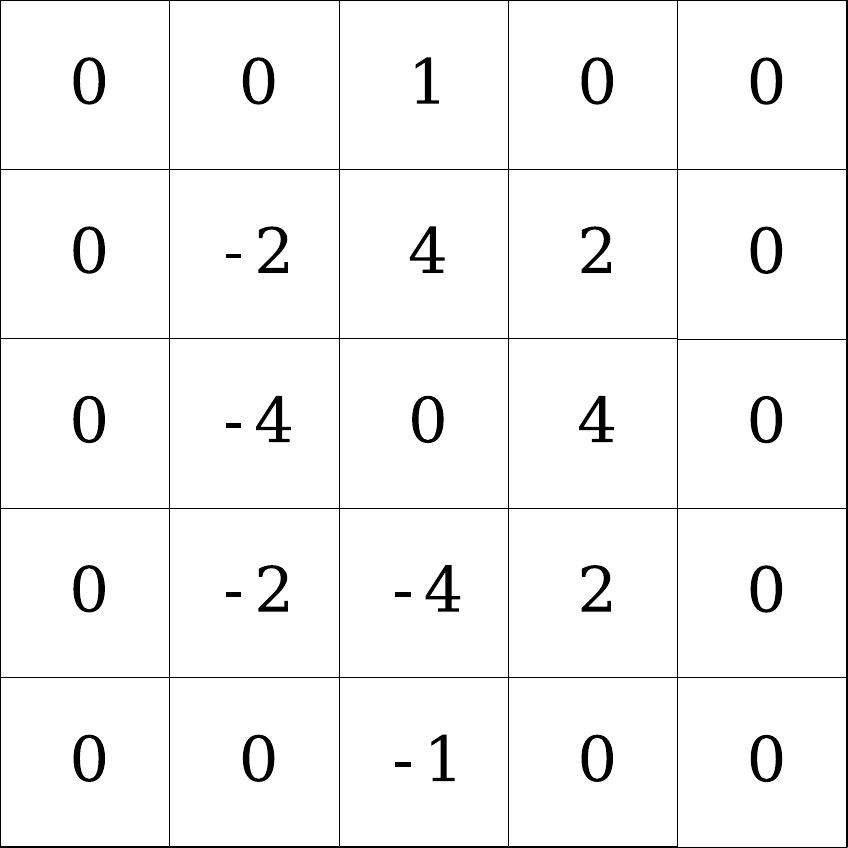}}
   \subfigure[]{\includegraphics[width=2.5cm]{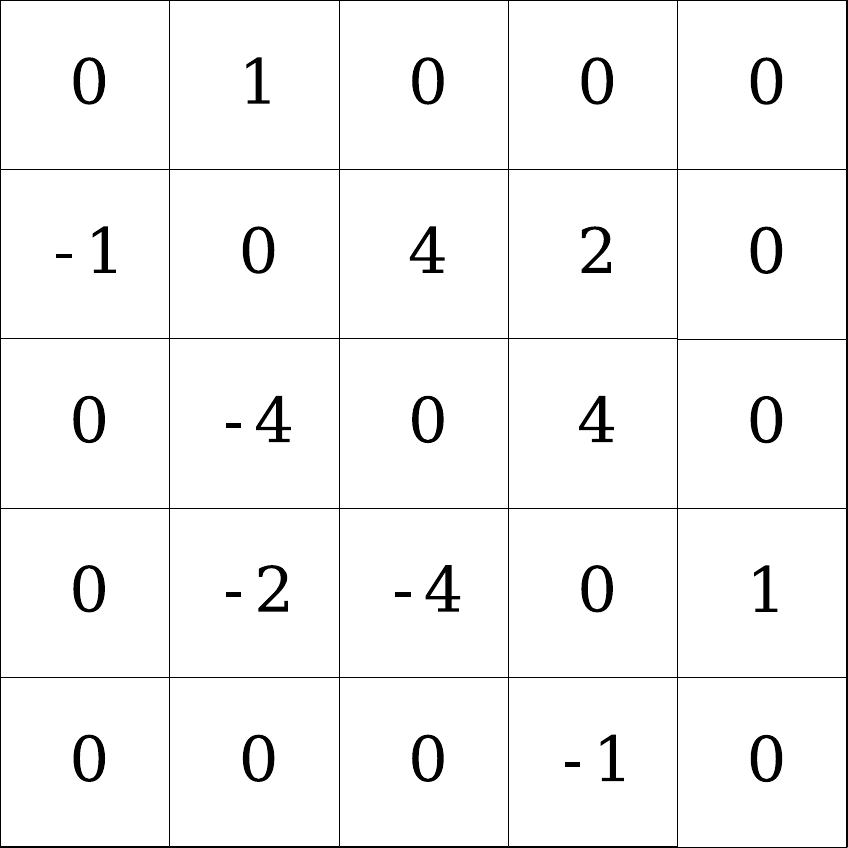}}
   \subfigure[]{\includegraphics[width=2.5cm]{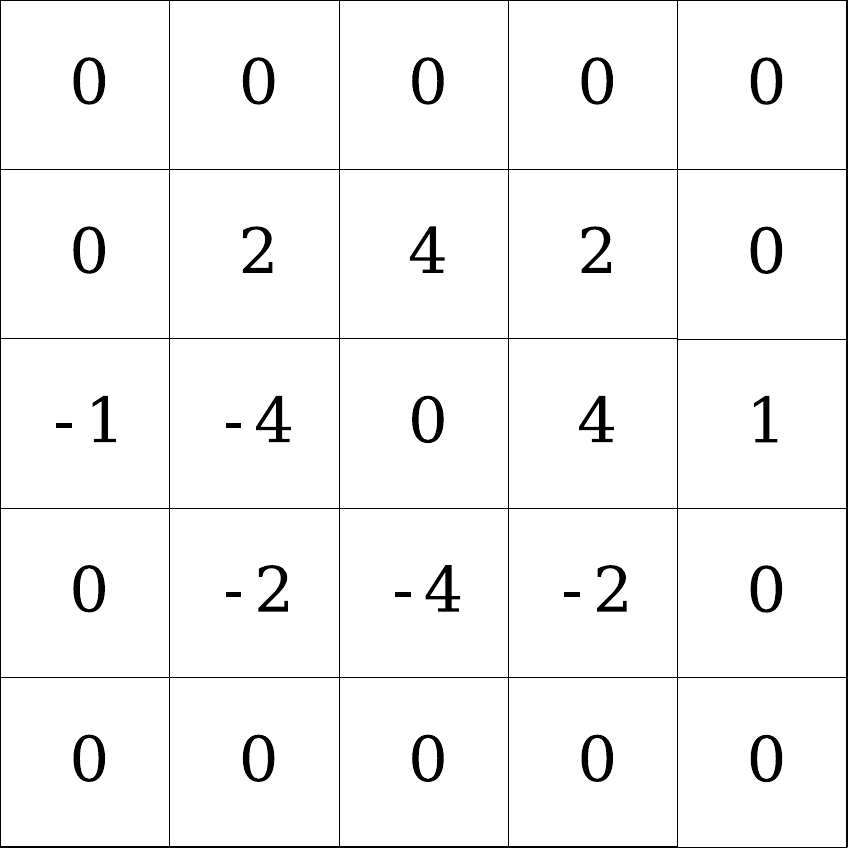}}
    \caption{Eight masks of eight-direction Sobel algorithm}
    \label{fig3}
\end{figure}

Comparing the gradient values to the threshold, this pixel will be considered part of the edge when $G \ge T$.

\section{Quantum image edge detection based on the eight-direction Sobel operator}\label{Section 3}

\subsection{Quantum operations}\label{subsec2}

\begin{enumerate}[(1)]

\item Quantum comparator 

Quantum comparator (QC) \cite{Oliveira2007} can compare the magnitude relationship between two numbers. 
 It takes two $n$ qubits sequences $\lvert A \rangle  = \lvert {{a_{n - 1}}{a_{n - 2}} \cdots {a_1}{a_0}} \rangle $ and $\lvert B \rangle  = \lvert {{b_{n - 1}}{b_{n - 2}} \cdots {b_1}{b_0}} \rangle $ as the input and $C_1$, $C_0$ as the output. If $A > B$, then ${C_1} = 1$ and ${C_0} = 0$; if $A<B$, then ${C_1} = 0$ and ${C_0} = 1$; if $A=B$, then ${C_1} = 0$ and ${C_0} = 0$. A schematic diagram of the  QC   is shown in Fig. \ref{Fig4}.
\begin{figure}
    \centering
    \includegraphics[width=6cm]{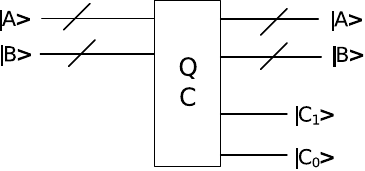}
    \caption{Quantum circuit realization of QC}
    \label{Fig4}
\end{figure}
\item  Cycle shift transformation

The cyclic shift transformation operation (CT) \cite{Le2010,Le2011S,Wang2014}  is moving all pixels in an image simultaneously several units in the $X$ or $Y$ direction. For $n$ qubits sequence $\lvert Y \rangle  = \lvert {{y_{n - 1}}{y_{n - 2}} \cdots {y_1}{y_0}} \rangle $, the CT operation can implement  ${2^n} + 1$ and ${2^n} - 1$.  Fig. \ref{Fig5} shows the schematic diagram  of the CT operation: $CT(+1)$ and $CT(-1)$. They are $\lvert {(Y + 1)\bmod {2^n}} \rangle $  and $\lvert {(Y - 1)\bmod {2^n}} \rangle $.
\begin{figure}
    \centering
    \subfigure[]{\includegraphics[width=5.5cm]{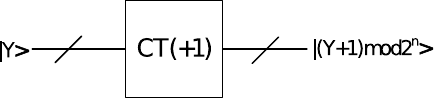}}
    \subfigure[]{\includegraphics[width=5.5cm]{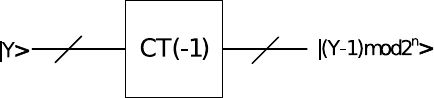}}
    \caption{Diagram of CT($+1$) and CT($-1$) operation}
    \label{Fig5}
\end{figure}
\item Reversible parallel adder

The reversible parallel adder (PA)  \cite{Islam2009} can compute the addition of $n$ qubits sequence $\lvert A \rangle  = \lvert {{a_{n - 1}}{a_{n - 2}} \cdots {a_1}{a_0}} \rangle $ and $n$ qubits sequence $\lvert B \rangle  = \lvert {{b_{n - 1}}{b_{n - 2}} \cdots {b_1}{b_0}} \rangle $.  It takes  $\lvert A \rangle  $ and $\lvert B \rangle $ as the input and $\lvert A+B \rangle$ as the output.  As shown in Fig. \ref{Fig6}.
\begin{figure}
    \centering
    \includegraphics[width=6cm]{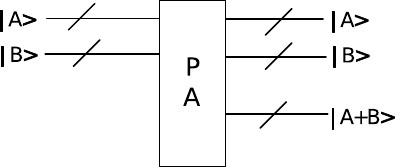}
    \caption{Quantum circuit realization of PA operation}
    \label{Fig6}
\end{figure}

\item Quantum absolute value operation

The quantum absolute value operation is used to calculate the absolute value of two integers in a quantum circuit, and subtraction of a binary bit sequence can be converted to the addition of complement. Quantum subtractor circuits can therefore be designed through a combination of quantum PA operation and complement  operation (CA) \cite{Iliyasu2013,Thapliyal2009,Thapliyal2011}. Assuming that $x = {x_n}{x_{n - 1}} \cdots {x_1}{x_0}$  is a signed binary integer, the highest bit is symbolic bit (0 represents the $x$ as a positive number and 1 represents the $x$ as a negative number) and the other bits represent value. The complement operation for the binary number $x$ is \cite{Li2018}:
$${\left[ x \right]_{{\rm{CA}}}} = \left\{ {\begin{array}{*{20}{c}}
{0,{x_{n - 1}}{x_{n - 2}} \cdots {x_1}{x_0},{x_n} = 0}\\
{1,\overline {{x_{n - 1}}} \overline {{x_{n - 2}}}  \cdots \overline {{x_1}} \overline {{x_0}} ,{x_n} = 1}
\end{array}} \right.\eqno{(6)}$$
Among them, $\overline {{x_k}}  = 1 - {x_k}, k = 0,1, \cdots ,n - 1$. The complement operation is shown in Fig. \ref{Fig7}.
\begin{figure}
    \centering
    \includegraphics[width=11cm]{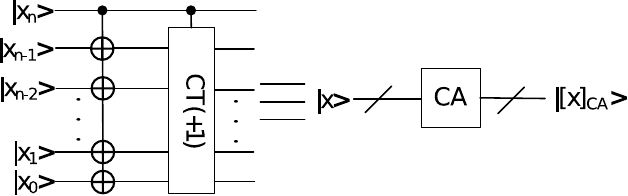}
    \caption{Quantum circuit realization of CA}
    \label{Fig7}
\end{figure}

Thus, the subtraction operation of computing two integers can be written as :$$A - B = A + {( - B)_{\rm{CA}}} = {\left[ {A + (\overline B  + 1)} \right]_{\rm{CA}}}\eqno{(7)}$$
Among them, $\overline B  = {b_n}\overline {{b_{n - 1}}} \overline {{b_{n - 2}}}  \cdots \overline {{b_1}} \overline {{b_0}} $. Suppose the value of $A-B$ is expressed as a $n+ 1$ bits binary number with a signed bit : $ D = {d_n}{d_{n - 1}}{d_{n - 2}} \cdots {d_1}{d_0}$, where the $d_n$ is a sign  bit. So while ignoring the sign bit, the absolute value of $A-B$ is ${d_n}{d_{n - 1}}{d_{n - 2}} \cdots {d_1}{d_0}$. Therefore, the quantum circuit for calculating the absolute value of the $\lvert{A-B}\rvert$ operation is shown in Fig. \ref{Fig8}.
\begin{figure}
    \centering
    \includegraphics[width=11cm]{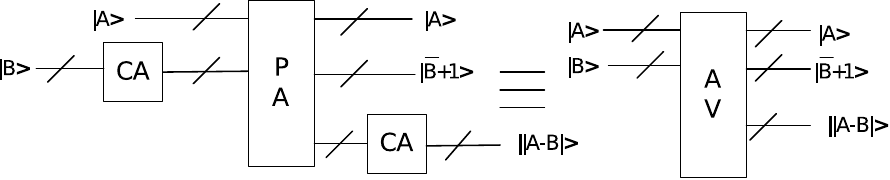}
    \caption{Quantum circuit of AV operation}
    \label{Fig8}
\end{figure}

\item Quantum double operation 

The quantum double operation (DO) \cite{Chetia2021,Li2018B} is used to multiply an integer binary bits by 2. The quantum circuit of this quantum operation based on quantum Swap gates operation and auxiliary qubits is shown in Fig. \ref{Fig9}.
\begin{figure}
    \centering
    \includegraphics[width=11cm]{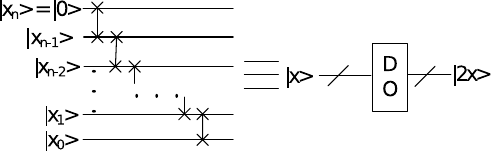}
    \caption{Quantum circuit realization of DO}
    \label{Fig9}
\end{figure}
\item Quantum copy operation

 The quantum copy operstion is completed with quantum controlled not-gates (CNOT) and auxiliary qubits \cite{Iliyasu2013}. The quantum circuit is shown in Fig. \ref{Fig10}.

 \begin{figure}
    \centering
    \includegraphics[width=11cm]{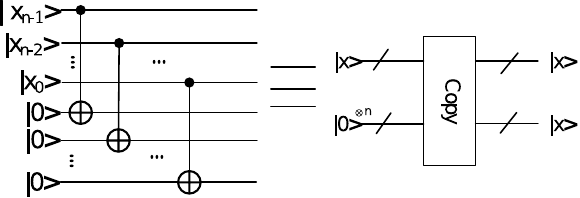}
    \caption{ Quantum circuit realization of Copy}
    \label{Fig10}
\end{figure}
\end{enumerate}
 \subsection{Quantum circuit realization for edge detection}
In this subsection, we  introduce the workflow of the whole edge detection algorithm first. And then,  the corresponding quantum circuits according to the workflow are designed.
    
\begin{figure}
    \centering
    \includegraphics[width=11.5cm]{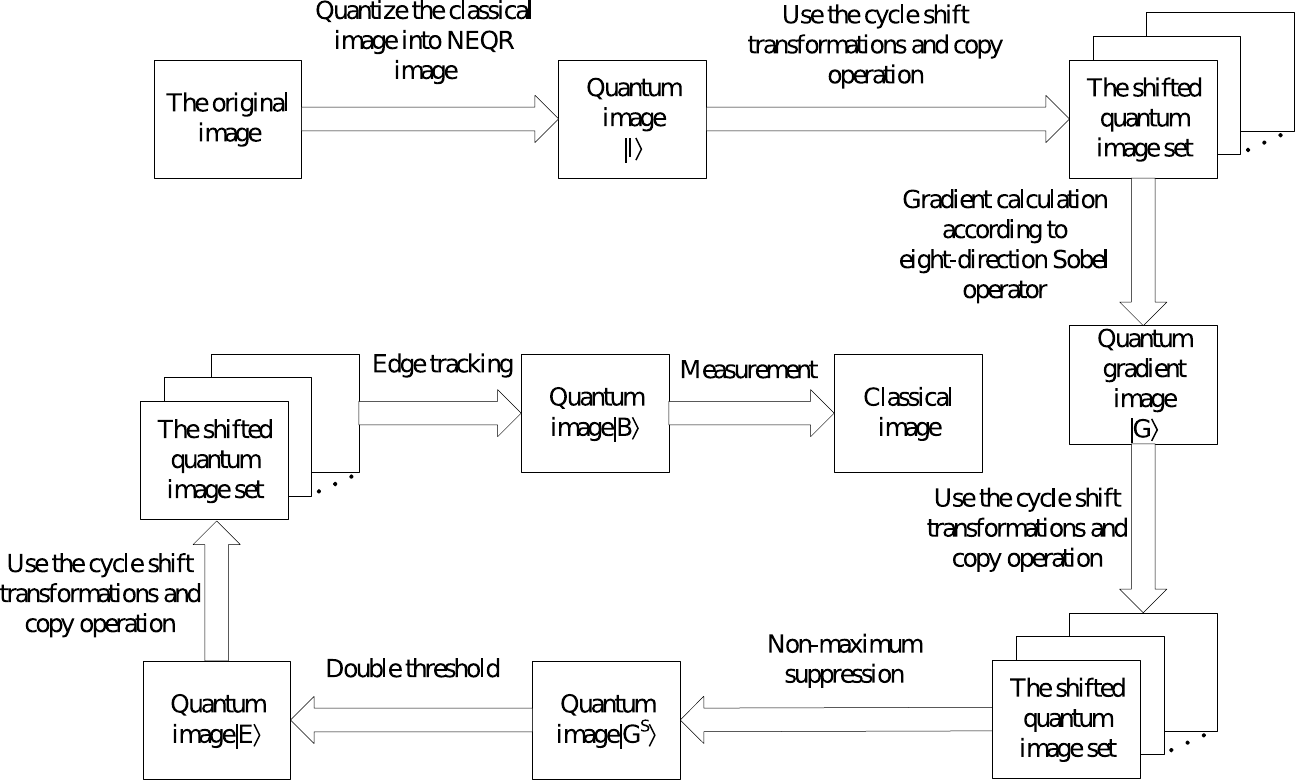}
    \caption{Workflow of our proposed algorithm}
    \label{Fig11}
\end{figure}

Figure \ref{Fig11} represents the workflow of the quantum image edge detection algorithm based on eight-direction Sobel operator, mainly consisting of six steps ---  quantum image preparation, quantum image set shift transformation, quantum image gradient value ccalculation, non-maximum suppression, double threshold detection and edge tracking.
The original image is represented as a NEQR  image firstly. Then, according to the Sobel operator, the quantum images obtained in the first step are cycle shift-transformed. Following that,  the gradient $\lvert G \rangle $ for each pixel is calculated by using the  Sobel operator. Then, each pixel is processed with non-maximum suppression to eliminate edge false detection and stored as the maximum gradient $\lvert {{G^s}} \rangle $. In addition, the gradient values of all pixels are compared with the double threshold to obtain strong and weak edges $\lvert E \rangle $. Finally, edge tracking is used to obtain the true edge $\lvert B \rangle $. 
\\
\noindent\textbf{Step 1 NEQR  images  preparation.} In order to turn a digital image into a quantum image, (${2n+q}$) qubits are required to store  ${ 2^n\times2^n}$ size of an image. Furthermore, 24 extra  qubits are required to record the   color information of the shifted pixels in next step, which can be prepared by using the tensor product of auxiliary qubits and quantum image $\lvert I \rangle $, i.e.,

$$\begin{array}{l}
{\lvert 0 \rangle ^{ \otimes 24q}} \otimes \lvert {{I_{YX}}} \rangle  = \frac{1}{{{2^n}}}\sum\limits_{Y = 0}^{{2^n} - 1} {\sum\limits_{X = 0}^{{2^n} - 1} {{{\lvert 0 \rangle }^{ \otimes 24q}}\lvert {{C_{YX}}} \rangle } } \lvert Y \rangle \lvert X \rangle \\
\begin{array}{*{20}{c}}
{}&{}&{}&{}&{}&{}&{}&{}&{}&{}&{}&{}&{}&{\begin{array}{*{20}{c}}
{}&{}& {}&{}&= 
\end{array}}
\end{array}\frac{1}{{{2^n}}}\sum\limits_{Y = 0}^{{2^n} - 1} {\sum\limits_{X = 0}^{{2^n} - 1} {{{\lvert 0 \rangle }^{ \otimes q}}} }  \cdots {\lvert 0 \rangle ^{ \otimes q}}\lvert Y \rangle \lvert X \rangle 
\end{array}\eqno{(8)}$$ 
\noindent\textbf{Step 2 Quantum image set shift transformation.} Following the steps in Table \ref{tab1}, the neighborhood pixels of the entire image $\lvert{I_{Y\!X}}\rangle$ are acquired and stored in additional qubits. In this step, every time a shift operation is performed, we use the Copy operation to copy the gray-scale value information of the shifted pixels into the prepared qubits to get 24 quantum images, and the pixels in the 24  quantum images  are simultaneously shifted within the $5\times5$ neighborhood pixels using CT operation.  Specific quantum operations of any  $5\times5$ neighborhood pixels are as follows:
$$
    \begin{aligned}
    \begin{array}{l}
\frac{1}{{{2^n}}}\sum\limits_{Y = 0}^{{2^n} - 1} {\sum\limits_{X = 0}^{{2^n} - 1} {\lvert {{C_{Y - 2,X - 2}}} \rangle  \otimes \lvert {{C_{Y - 1,X - 2}}} \rangle  \otimes \lvert {{C_{Y,X - 2}}} \rangle  \otimes \lvert {{C_{Y + 1,X - 2}}} \rangle   } } \\
\begin{array}{*{20}{c}}
{}&{}& \otimes \lvert {{C_{Y + 2,X - 2}}} \rangle\otimes 
\end{array}\lvert {{C_{Y - 2,X - 1}}} \rangle  \otimes \lvert {{C_{Y - 1,X - 1}}} \rangle  \otimes \lvert {{C_{Y,X - 1}}} \rangle  \otimes \lvert {{C_{Y + 1,X - 1}}} \rangle   \\
\begin{array}{*{20}{c}}
{}&{}& \otimes \lvert {{C_{Y + 2,X - 1}}} \rangle\otimes 
\end{array}\lvert {{C_{Y - 2,X}}} \rangle  \otimes \lvert {{C_{Y - 1,X}}} \rangle  \otimes \lvert {{C_{Y,X}}} \rangle  \otimes \lvert {{C_{Y + 1,X}}} \rangle  \otimes \lvert {{C_{Y + 2,X}}} \rangle \\
\begin{array}{*{20}{c}}
{}&{}& \otimes 
\end{array}\lvert {{C_{Y - 2,X + 1}}} \rangle  \otimes \lvert{{C_{Y - 1,X + 1}}} \rangle  \otimes \lvert {{C_{Y,X + 1}}}\rangle  \otimes \lvert {{C_{Y + 1,X + 1}}} \rangle  \otimes \lvert {{C_{Y + 2,X + 1}}} \rangle \\
\begin{array}{*{20}{c}}
{}&{}& \otimes 
\end{array}\lvert {{C_{Y - 2,X + 2}}} \rangle  \otimes \lvert {{C_{Y - 1,X + 2}}} \rangle  \otimes \lvert {{C_{Y,X + 2}}} \rangle  \otimes \lvert {{C_{Y + 1,X + 2}}} \rangle  \otimes \lvert{{C_{Y + 2,X + 2}}} \rangle \\ 
\begin{array}{*{20}{c}}
{}&{}& \otimes 
\end{array} \lvert Y \rangle \lvert X \rangle 
\end{array}
    \end{aligned}
    \eqno{(9)}
$$
\begin{tableorg}
\begin{minipage}{338pt}
\caption{Computation prepared algorithm for shifting the image}\label{tab1}%
\resizebox{.99\columnwidth}{!}{
\begin{tabular}{@{}l@{}}
\midrule
1. Input: the original NEQR image ${I_{YX}}$,$ \lvert {{I_{YX}}} \rangle  = \frac{1}{{{2^n}}}\sum\limits_{Y = 0}^{{2^n} - 1} {\sum\limits_{X = 0}^{{2^n} - 1} {\lvert {{C_{YX}}} \rangle } } \lvert Y \rangle \lvert X \rangle $\\  
2. Shift ${I_{YX}}$one unit upward, then$\lvert  {{I_{Y + 1X}}} \rangle  = CT\lvert ( {Y - } )\lvert  {{I_{YX}}} \rangle  = \frac{1}{{{2^n}}}\!\sum\limits_{Y = 0}^{{2^n} - 1} \!{\sum\limits_{X = 0}^{{2^n} - 1} {\lvert  {{C_{Y + 1X}}} \rangle } } \lvert  Y \rangle \lvert  X \rangle $\\
3. Shift${I_{Y + 1X}}$one unit leftward, then$\lvert {{I_{Y +1X + 1}}} \rangle  = CT\left( {X - } \right)\lvert {{I_{Y + 1X}}} \rangle  = \!\frac{1}{{{2^n}}}\!\sum\limits_{Y= 0}^{{2^n} - 1} {\sum\limits_{X = 0}^{{2^n} - 1} {\lvert {{C_{Y + 1X + 1}}} \rangle } } \lvert Y \rangle \lvert X \rangle $\\
4. Shift${I_{Y + 1X + 1}}$one unit downward, then$\lvert {{I_{YX + 1}}} \rangle  = CT\left( {Y + } \right)\lvert {{I_{Y + 1X + 1}}} \rangle  = \frac{1}{{{2^n}}}\sum\limits_{Y = 0}^{{2^n} - 1} {\sum\limits_{X = 0}^{{2^n} - 1} {\lvert {{C_{YX + 1}}} \rangle } } \lvert Y \rangle \lvert X \rangle $\\
5. Shift${I_{YX + 1}}$one unit downward, then$\lvert {{I_{Y - 1X + 1}}} \rangle  = CT\left( {Y + } \right)\lvert {{I_{YX + 1}}} \rangle  = \frac{1}{{{2^n}}}\sum\limits_{Y = 0}^{{2^n} - 1} {\sum\limits_{X = 0}^{{2^n} - 1} {\lvert {{C_{Y - 1X + 1}}} \rangle } } \lvert Y \rangle \lvert X \rangle $\\
6. Shift${I_{Y - 1X + 1}}$one unit rightward, then$ \lvert {{I_{Y - 1X}}} \rangle  = CT\left( {X + } \right)\lvert {{I_{Y - 1X + 1}}} \rangle  = \frac{1}{{{2^n}}}\sum\limits_{Y = 0}^{{2^n} - 1} {\sum\limits_{X = 0}^{{2^n} - 1} {\lvert  {{C_{Y - 1X}}} \rangle } } \lvert  Y \rangle \lvert  X \rangle $\\
7. Shift${I_{Y - 1X}}$one unit rightward, then$\lvert {{I_{Y - 1X - 1}}} \rangle  = CT\left( {X + } \right)\lvert {{I_{Y - 1X}}} \rangle  = \frac{1}{{{2^n}}}\sum\limits_{Y = 0}^{{2^n} - 1} {\sum\limits_{X = 0}^{{2^n} - 1} {\lvert {{C_{Y - 1X - 1}}} \rangle } } \lvert Y \rangle \lvert X \rangle $\\
8. Shift${I_{Y - 1X-1}}$one unit upward, then$\lvert {{I_{YX - 1}}} \rangle  = CT\left( {Y - } \right)\lvert {{I_{Y - 1X - 1}}} \rangle  = \frac{1}{{{2^n}}}\sum\limits_{Y = 0}^{{2^n} - 1} {\sum\limits_{X = 0}^{{2^n} - 1} {\lvert {{C_{YX - 1}}} \rangle } } \lvert Y \rangle \lvert X \rangle$\\
9. Shift${I_{YX-1}}$one unit upward, then$\lvert {{I_{Y + 1X - 1}}}\rangle  = CT\left( {Y - } \right)\lvert {{I_{YX - 1}}} \rangle  = \frac{1}{{{2^n}}}\sum\limits_{Y = 0}^{{2^n} - 1} {\sum\limits_{X = 0}^{{2^n} - 1} {\lvert {{C_{Y + 1X - 1}}} \rangle } } \lvert Y \rangle \lvert X \rangle $\\
10. Shift${I_{Y+1X-1}}$one unit upward, then$\lvert {{I_{Y + 2X - 1}}}\rangle  = CT\left( {Y - } \right)\lvert {{I_{Y + 1X - 1}}} \rangle  = \frac{1}{{{2^n}}}\sum\limits_{Y = 0}^{{2^n} - 1} {\sum\limits_{X = 0}^{{2^n} - 1} {\lvert {{C_{Y + 2X - 1}}} \rangle } } \lvert Y \rangle \lvert X \rangle  $\\
11. Shift${I_{Y+2X-1}}$one unit leftward, then$\lvert {{I_{Y + 2X}}} \rangle  = CT\left( {X - } \right)\lvert {{I_{Y + 2X - 1}}} \rangle  = \frac{1}{{{2^n}}}\sum\limits_{Y = 0}^{{2^n} - 1} {\sum\limits_{X = 0}^{{2^n} - 1} {\lvert {{C_{Y + 2X}}} \rangle } } \lvert Y \rangle \lvert X \rangle $\\
12. Shift${I_{Y+2X}}$one unit leftward, then$\lvert {{I_{Y + 2X + 1}}} \rangle  = CT\left( {X - } \right)\lvert {{I_{Y + 2X}}} \rangle  = \frac{1}{{{2^n}}}\sum\limits_{Y = 0}^{{2^n} - 1} {\sum\limits_{X = 0}^{{2^n} - 1} {\lvert {{C_{Y + 2X + 1}}} \rangle } } \lvert Y\rangle \lvert X \rangle $\\
13. Shift${I_{Y+2X+1}}$one unit leftward, then$\lvert {{I_{Y + 2X + 2}}} \rangle  = CT\left( {X - } \right)\lvert{{I_{Y + 2X + 1}}} \rangle  = \frac{1}{{{2^n}}}\sum\limits_{Y = 0}^{{2^n} - 1} {\sum\limits_{X = 0}^{{2^n} - 1} {\lvert {{C_{Y + 2X + 2}}} \rangle } } \lvert Y \rangle \lvert X \rangle $\\
14. Shift${I_{Y+2X+2}}$one unit downward, then$\lvert {{I_{Y + 1X + 2}}} \rangle  = CT\left( {Y + } \right)\lvert {{I_{Y + 2X + 2}}} \rangle  = \frac{1}{{{2^n}}}\sum\limits_{Y = 0}^{{2^n} - 1} {\sum\limits_{X = 0}^{{2^n} - 1} {\lvert {{C_{Y + 1X + 2}}} \rangle } } \lvert Y \rangle \lvert X \rangle $\\
15. Shift ${I_{Y+1X+2}}$ one unit downward, then $\lvert {{I_{YX + 2}}} \rangle  = CT\left( {Y + } \right)\lvert {{I_{Y + 1X + 2}}} \rangle  = \frac{1}{{{2^n}}}\sum\limits_{Y = 0}^{{2^n} - 1} {\sum\limits_{X = 0}^{{2^n} - 1} {\lvert {{C_{YX + 2}}} \rangle } } \lvert Y\rangle \lvert X \rangle $\\
16. Shift ${I_{YX+2}}$ one unit downward, then $\lvert {{I_{Y - 1X + 2}}} \rangle  = CT\left( {Y + } \right)\lvert {{I_{YX + 2}}} \rangle  = \frac{1}{{{2^n}}}\sum\limits_{Y = 0}^{{2^n} - 1} {\sum\limits_{X = 0}^{{2^n} - 1} {\lvert{{C_{Y - 1X + 2}}} \rangle } } \lvert Y \rangle \lvert X \rangle$ \\
17. Shift${I_{Y-1X+2}}$ one unit downward, then$\lvert{{I_{Y - 2X + 2}}} \rangle  = CT\left( {Y + } \right)\lvert {{I_{Y - 1X + 2}}} \rangle  = \frac{1}{{{2^n}}}\sum\limits_{Y = 0}^{{2^n} - 1} {\sum\limits_{X = 0}^{{2^n} - 1} {\lvert {{C_{Y - 2X + 2}}} \rangle } } \lvert Y \rangle \lvert X \rangle $\\
18. Shift${I_{Y-2X+2}}$ one unit rightward, then$\vert {{I_{Y - 2X + 1}}} \rangle  = CT\left( {X + } \right)\vert {{I_{Y - 2X + 2}}} \rangle  = \frac{1}{{{2^n}}}\sum\limits_{Y = 0}^{{2^n} - 1} {\sum\limits_{X = 0}^{{2^n} - 1} {\vert {{C_{Y - 2X + 1}}} \rangle } } \vert Y \rangle \vert X \rangle $\\
19. Shift${I_{Y-2X+1}}$one unit rightward, then$\lvert {{I_{Y - 2X}}} \rangle  = CT\left( {X + } \right)\lvert {{I_{Y - 2X + 1}}} \rangle  = \frac{1}{{{2^n}}}\sum\limits_{Y = 0}^{{2^n} - 1} {\sum\limits_{X = 0}^{{2^n} - 1} {\lvert {{C_{Y - 2X}}} \rangle } } \lvert Y \rangle \lvert X \rangle$\\
20. Shift${I_{Y-2X}}$ oneunit rightward, then$\lvert {{I_{Y - 2X - 1}}} \rangle  = CT\left( {X + } \right)\lvert {{I_{Y - 2X}}} \rangle  = \frac{1}{{{2^n}}}\sum\limits_{Y = 0}^{{2^n} - 1} {\sum\limits_{X = 0}^{{2^n} - 1} {\lvert {{C_{Y - 2X - 1}}} \rangle } } \lvert Y \rangle \lvert X \rangle $\\
21. Shift${I_{Y-2X-1}}$one unit rightward, then$\lvert {{I_{Y - 2X - 2}}} \rangle  = CT\left( {X + } \right)\lvert {{I_{Y - 2X - 1}}} \rangle  = \frac{1}{{{2^n}}}\sum\limits_{Y = 0}^{{2^n} - 1} {\sum\limits_{X = 0}^{{2^n} - 1} {\lvert {{C_{Y - 2X - 2}}} \rangle } } \lvert Y \rangle \lvert X \rangle$\\
22. Shift${I_{Y-2X-2}}$one unit upward, then$\lvert {{I_{Y - 1X - 2}}} \rangle  = CT\left( {Y - } \right)\lvert {{I_{Y - 2X - 2}}} \rangle  = \frac{1}{{{2^n}}}\sum\limits_{Y = 0}^{{2^n} - 1} {\sum\limits_{X = 0}^{{2^n} - 1} {\lvert {{C_{Y - 1X - 2}}} \rangle } } \lvert Y \rangle \lvert X \rangle $\\
23. Shift${I_{Y-1X-2}}$one unit upward, then$\lvert {{I_{YX - 2}}} \rangle  = CT\left( {Y - } \right)\lvert {{I_{Y - 1X - 2}}} \rangle  = \frac{1}{{{2^n}}}\sum\limits_{Y = 0}^{{2^n} - 1} {\sum\limits_{X = 0}^{{2^n} - 1} {\lvert {{C_{YX - 2}}} \rangle } } \lvert Y \rangle \lvert X \rangle$\\
24. Shift${I_{YX-2}}$one unit upward, then$\lvert {{I_{Y + 1X - 2}}} \rangle  = CT\left( {Y - } \right)\lvert {{I_{YX - 2}}} \rangle  = \frac{1}{{{2^n}}}\sum\limits_{Y = 0}^{{2^n} - 1} {\sum\limits_{X = 0}^{{2^n} - 1} {\lvert {{C_{Y + 1X - 2}}} \rangle } } \lvert Y \rangle \lvert X \rangle $\\
25. Shift${I_{Y+1X-2}}$one unit upward, then$\lvert {{I_{Y + 2X - 2}}} \rangle  = CT\left( {Y - } \right)\lvert {{I_{Y + 1X - 2}}} \rangle  = \frac{1}{{{2^n}}}\sum\limits_{Y = 0}^{{2^n} - 1} {\sum\limits_{X = 0}^{{2^n} - 1} {\lvert {{C_{Y + 2X - 2}}} \rangle } } \lvert Y \rangle \lvert X \rangle $\\
26. Shift${I_{Y+2X-2}}$two units leftwards and two units downwards to the original position, then\\
\qquad$\lvert {{I_{YX}}} \rangle  = CT\left( {X - } \right)CT\left( {X - } \right)CT\left( {Y + } \right)CT\left( {Y + } \right)\lvert {{I_{Y + 1X - 2}}} \rangle  = \frac{1}{{{2^n}}}\sum\limits_{Y = 0}^{{2^n} - 1} {\sum\limits_{X = 0}^{{2^n} - 1} {\lvert {{C_{Y + 2X - 2}}} \rangle } } \lvert Y \rangle \lvert X \rangle $\\
\botrule
\end{tabular}
}
\end{minipage}
\end{tableorg}

\noindent\textbf{Step 3 Gradients calculation}. The gradients of the $\lvert {I_{Y\!X}}\rangle$ are calculated using the Sobel operator in eight directions. The specific calculations operation are as follows:
\begin{small}
\[\lvert {\lvert{G_{Y\!X}^0} \rangle } \rvert = \bigg\lvert \begin{array}{l}
\lvert {{C_{Y \!- \!2,X\! +\! 1}}} \rangle \! +\! \lvert{2{C_{Y \!-\! 1,X\! +\! 1}}} \rangle  +\! \lvert {4{C_{Y,X\! + 1\!}}} \rangle  \!+\! \lvert {2{C_{Y\! +\! 1,X\! +\! 1}}} \rangle \! +\! \lvert {{C_{Y\! +\! 2,X\! +\! 1}}} \rangle \\
 - \lvert{{C_{Y \!- \!2,X\! -\! 1}}} \rangle \! -\! \lvert {2{C_{Y \!- \!1,X \!- \!1}}} \rangle  - \!\lvert{4{C_{Y,X\! -\! 1}}} \rangle \! -\! \lvert {2{C_{Y \!+ \!1,X \!- \!1}}} \rangle \! -\! \lvert {{C_{Y \!+ \!2,X \!-\! 1}}} \rangle 
\end{array} \bigg\rvert\]
\[\lvert {\lvert {G_{Y\!X}^{22.5}} \rangle } \rvert = \bigg\lvert \begin{array}{l}
\lvert {{C_{Y \!+\! 2,X}}} \rangle \! +\! \lvert {2{C_{Y \!+ \!1,X\! + \!1}}} \rangle\!  + \!\lvert {2{C_{Y \!-\! 1,X\! +\! 1}}} \rangle \! + \!\lvert {4{C_{Y,X + \!1}}} \rangle \! +\! \lvert {4{C_{Y\! +\! 1,X}}} \rangle \\
 - \lvert {{C_{Y \!- \!2,X}}} \rangle \! - \!\lvert {2{C_{Y \!+\! 1,X\! -\! 1}}} \rangle \! -\! \lvert {2{C_{Y\! -\! 1,X \!- \!1}}} \rangle \! - \!\lvert{4{C_{Y,X \!-\! 1}}} \rangle \! - \!\lvert {4{C_{Y\! -\! 1,X}}} \rangle 
\end{array} \bigg\rvert\]
\[\lvert {\lvert {G_{Y\!X}^{45}} \rangle } \rvert = \bigg\lvert \begin{array}{l}
\lvert {{C_{Y \!+\! 2,X \!-\! 1}}} \rangle \! + \!\lvert {{C_{Y\! -\! 1,X\! + \!2}}} \rangle \! +\! \lvert {2{C_{Y\! + \!1,X \!+\! 1}}} \rangle \! + \!\lvert {4{C_{Y \!+ \!1,X}}} \rangle \! +\! \lvert {4{C_{Y,X \!+ \!1}}} \rangle \\
 - \lvert {{C_{Y\! +\! 1,X\! -\! 2}}} \rangle \! -\! \lvert {{C_{Y \!-\! 2,X \!+\! 1}}}\rangle  \!- \!\lvert {2{C_{Y \!-\! 1,X\! - \!1}}} \rangle \! - \!\lvert {4{C_{Y - 1,X}}} \rangle \! -\! \lvert {4{C_{Y,X\! -\! 1}}} \rangle 
\end{array} \bigg\rvert\]
\[\lvert {\lvert {G_{Y\!X}^{67.5}} \rangle } \lvert = \bigg\lvert \begin{array}{l}
\lvert {{C_{Y,X\! +\! 2}}}\rangle \! + \!\lvert {2{C_{Y \!+ \!1,X\! +\! 1}}} \rangle \! + \!\lvert {2{C_{Y\! + \!1,X \!- \!1}}} \rangle \! + \!\lvert {4{C_{Y \!+\! 1,X}}} \rangle \! +\! \lvert {4{C_{Y,X \!+\! 1}}} \rangle \\
 - \lvert {{C_{Y,X \!-\!2}}}\rangle \! -\! \lvert{2{C_{Y \!-\! 1,X \!+ \!1}}} \rangle \! - \!\lvert{2{C_{Y \!-\! 1,X\! -\! 1}}} \rangle \! - \!\lvert{4{C_{Y \!-\! 1,X}}} \rangle \! -\! \lvert {4{C_{Y,X\! -\! 1}}} \rangle 
\end{array} \bigg\rvert\]
\[\lvert {\lvert {G_{Y\!X}^{90}} \rangle } \rvert = \bigg\lvert \begin{array}{l}
\lvert {{C_{Y\! + \!1,X \!- \!2}}} \rangle \! + \!\lvert {{C_{Y \!+ \!1,X \!+ \!2}}} \rangle \! +\! \lvert {2{C_{Y \!+ \!1,X\! - \!1}}} \rangle \! +\! \lvert {2{C_{Y\! +\! 1,X \!+ \!1}}} \rangle \! +\! \lvert {4{C_{Y \!+\! 1,X}}} \rangle \\
 - \lvert {{C_{Y \!-\! 1,X \!-\! 2}}} \rangle \! - \!\lvert {{C_{Y \!-\! 1,X\! +\! 2}}} \rangle \! -\! \lvert {2{C_{Y \!-\! 1,X\! - \!1}}} \rangle \! -\! \lvert {2{C_{Y \!- \!1,X\! + \!1}}} \rangle \! -\!\lvert {4{C_{Y\! -\! 1,X}}} \rangle 
\end{array} \bigg\rvert\]
\[\lvert {\lvert {G_{Y\!X}^{112.5}}\rangle } \rvert = \bigg\lvert \begin{array}{l}
\lvert {{C_{Y,X \!- \!2}}} \rangle \! +\! \lvert{2{C_{Y\! +\! 1,X \!- \!1}}} \rangle \! + \!\lvert {2{C_{Y\! +\! 1,X \!+\! 1}}} \rangle \! + \!\lvert {4{C_{Y \!+ \!1,X}}} \rangle \! +\! \lvert {4{C_{Y,X\! - \!1}}}  \rangle \\
 - \lvert {{C_{Y,X\! +\! 2}}} \rangle \! - \!\lvert{2{C_{Y \!-\!1,X \!+\! 1}}} \rangle \! -\! \lvert {2{C_{Y\! - \!1,X \!-\! 1}}}  \rangle \! - \!\lvert {4{C_{Y \!-\! 1,X}}}  \rangle  \!- \!\lvert {4{C_{Y,X\! + \!1}}}  \rangle 
\end{array} \bigg\rvert\]
\[\lvert {\lvert {G_{Y\!X}^{135}} \rangle } \rvert = \bigg\lvert \begin{array}{l}
\lvert {{C_{Y \!-\! 1,X\! - \!2}}} \rangle \! +\! \lvert {{C_{Y \!+\! 1,X \!+ \!1}}} \rangle \! +\! \lvert {2{C_{Y\! +\! 1,X\! -\! 1}}} \rangle \! +\! \lvert {4{C_{Y\! +\! 1,X}}} \rangle \! +\! \lvert {4{C_{Y,X \!- \!1}}} \rangle \\
 - \lvert {{C_{Y\! - \!2,X \!- \!1}}} \rangle \! -\! \lvert {{C_{Y\! +\! 1,X\! +\! 2}}} \rangle  \!-\! \lvert {2{C_{Y \!-\! 1,X\! +\! 1}}} \rangle \! -\! \lvert {4{C_{Y \!-\! 1,X}}} \rangle \! -\! \lvert {4{C_{Y,X\! +\! 1}}} \rangle 
\end{array} \bigg\rvert\]
$$\lvert {\lvert{G_{Y\!X}^{157.5}} \rangle } \rvert =\bigg \lvert \begin{array}{l}
\lvert {{C_{Y\! +\! 2,X}}}  \rangle \! + \!\lvert {2{C_{Y\! +\! 1,X\! -\! 1}}} \rangle \! +\! \lvert {2{C_{Y\! - \!1,X\! - \!1}}} \rangle \! +\! \lvert {4{C_{Y\! + \!1,X}}} \rangle  \!+ \!\lvert {4{C_{Y,X \!- \!1}}}  \rangle \\
 - \lvert {{C_{Y \!- \!2,X}}} \rangle \! - \!\lvert{2{C_{Y\! +\! 1,X \!+\! 1}}} \rangle \! -\! \lvert{2{C_{Y\! - \!1,X\! +\! 1}}} \rangle \! - \!\lvert{4{C_{Y \!-\!1,X}}} \rangle  \!- \!\lvert {4{C_{Y,X\! +\! 1}}}  \rangle 
\end{array}\bigg \rvert\eqno{(10)}$$
\end{small}
Thus, the  gradient for each pixel is
\begin{small}
$$\lvert {\lvert G \rangle } \rvert = \max \left\{ \!{\lvert {\lvert {{G^0}} \rangle } \rvert,\lvert {\lvert {{G^{22.5}}} \rangle } \rvert,\lvert {\lvert {{G^{45}}} \rangle } \rvert,\lvert {\lvert {{G^{67.5}}} \rangle } \rvert,\lvert {\lvert {{G^{90}}} \rangle } \rvert,\lvert {\lvert {{G^{112.5}}} \rangle } \rvert,\lvert {\lvert {{G^{135}}} \rangle } \rvert,\lvert {\lvert {{G^{157.5}}} \rangle } \rvert}\! \right\}\eqno{(11)}
$$
\end{small}

Through quantum operations such as the quantum absolute value operation and the quantum comparator, the  gradient of each pixel can be obtained from Eq. (11). The gradient values $\lvert G \rangle$ can be expressed as:
$$\lvert G \rangle=\frac{1}{{{2^n}}}\sum\limits_{Y = 0}^{{2^n} - 1} {\sum\limits_{X = 0}^{{2^n} - 1} {\lvert N \rangle \lvert {G_{YX}^d} \rangle } } \lvert Y \rangle \lvert X \rangle \eqno{(12)}$$
where $d=$${0^ \circ }$, ${22.5^ \circ }$, ${45^ \circ }$, ${67.5^ \circ }$, ${90^ \circ }$, ${112.5^ \circ }$, ${135^ \circ }$, ${157.5^ \circ }$; $\lvert{N} \rangle  = \lvert 1\rangle$ for gradient values and $\lvert{N} \rangle  =\lvert 0\rangle$ for non-gradient values.

\begin{figure}
    \centering
    \includegraphics[width=9cm]{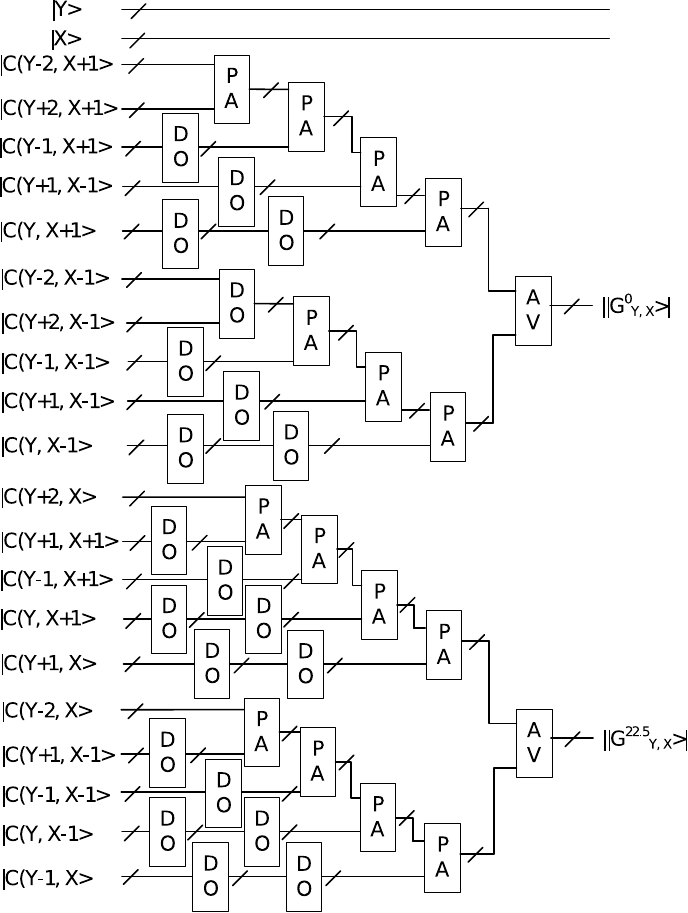}
    \caption{Quantum circuit realization for gradient   value calculation of a quantum image into the ${0^ \circ }$ and ${22.5^ \circ }$ directions}
    \label{Fig12}
\end{figure}

\begin{figure}
    \centering
    \includegraphics[width=9cm]{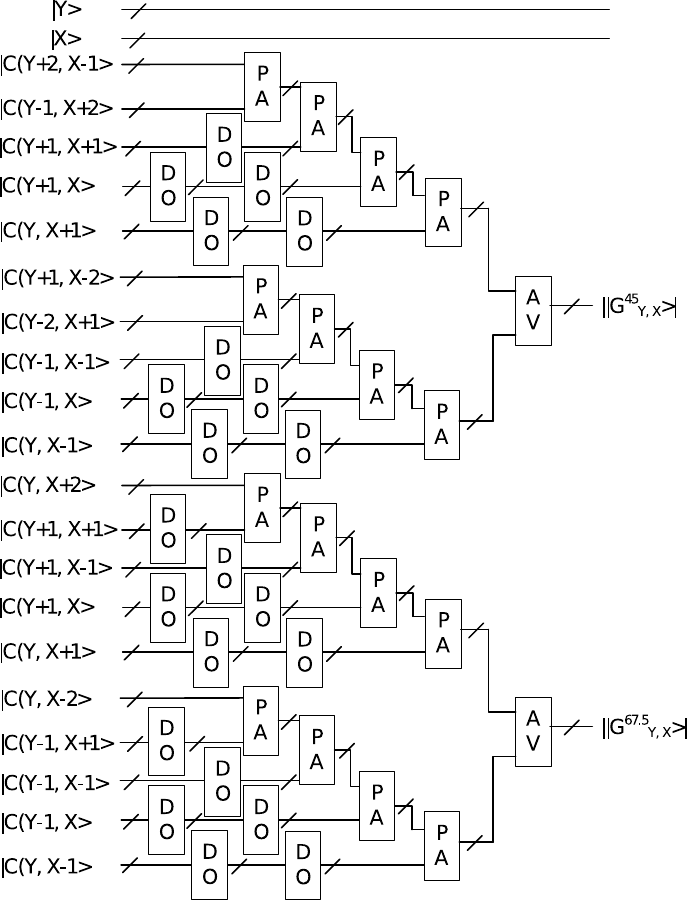}
    \caption{Quantum circuit realization for gradient   value calculation of a quantum image into the ${45^ \circ }$ and ${67.5^ \circ }$ directions}
    \label{Fig13}
\end{figure}

\begin{figure}
    \centering
    \includegraphics[width=9cm]{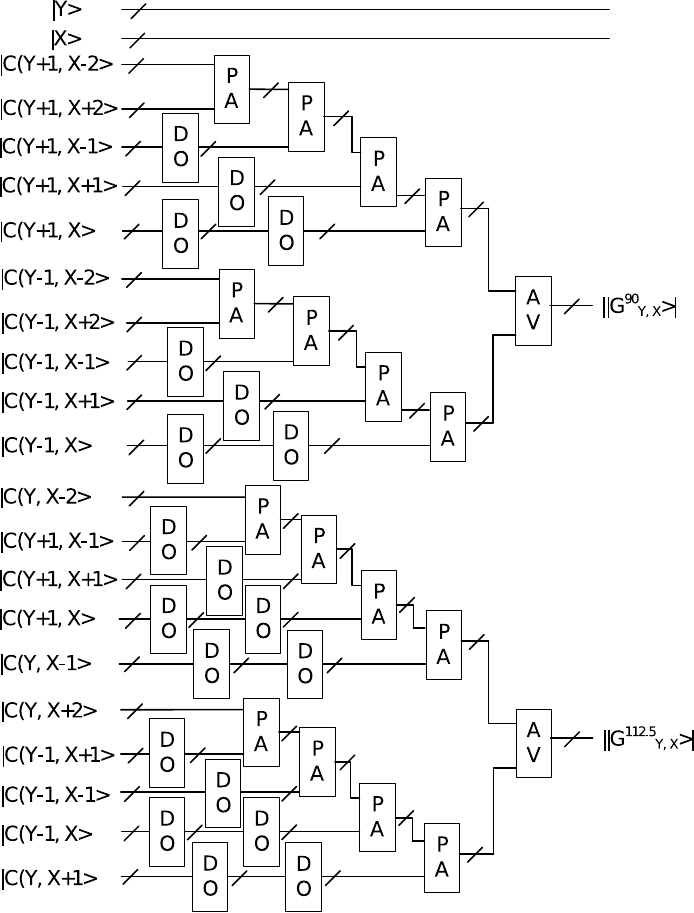}
    \caption{Quantum circuit realization for gradient   value calculation of a quantum image into the ${90^ \circ }$ and ${112.5^ \circ }$ directions}
    \label{Fig14}
\end{figure}

\begin{figure}
    \centering
    \includegraphics[width=9cm]{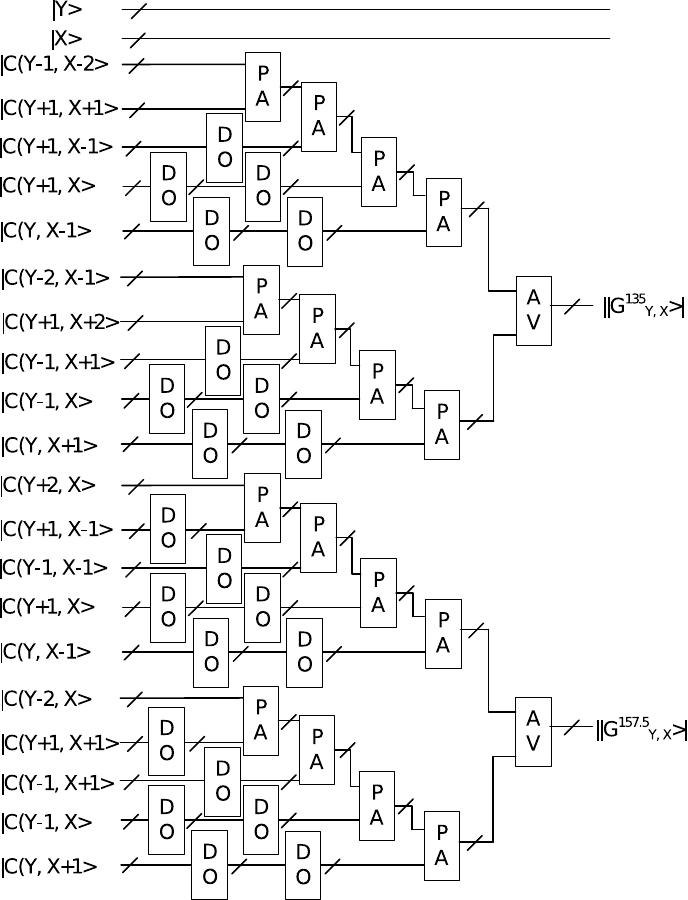}
    \caption{Quantum circuit realization for gradient   value calculation of a quantum image into the ${135^ \circ }$ and ${157.5^ \circ }$ directions}
    \label{Fig15}
\end{figure}

\begin{figure}
    \centering
    \includegraphics[width=11.5cm]{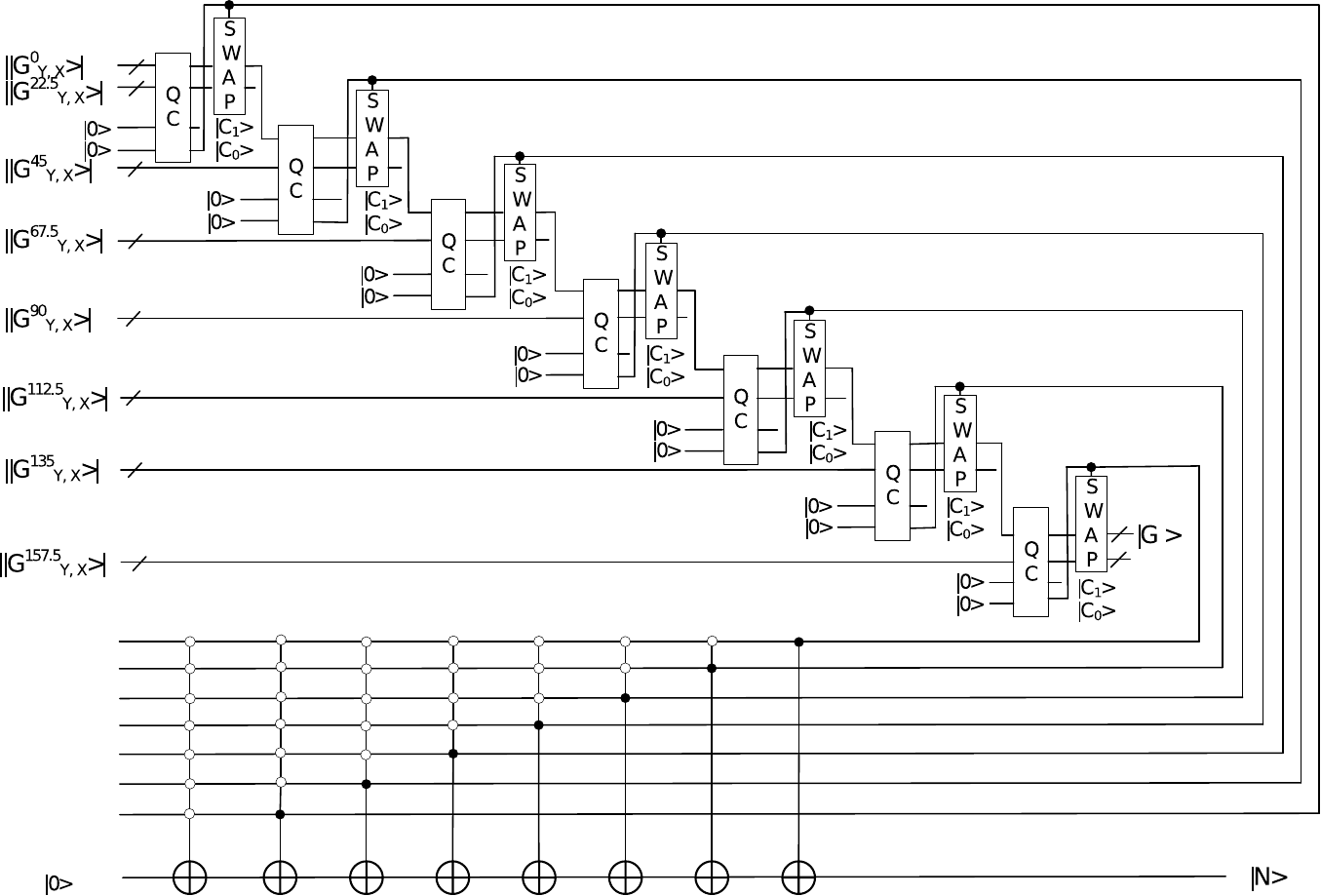}
    \caption{Quantum circuit realization of the  gradient calculation of image}
    \label{Fig16}
\end{figure}

Figures \ref{Fig12}-\ref{Fig15} show the gradient values calculation quantum circuits in eight directions. The quantum circuit for the  gradient calculation of the quantum image is shown in Fig. \ref{Fig16}. The oblique lines in the circuits represent n qubits, and the measurements and some auxiliary qubits are omitted.

\begin{figure}
    \centering
    \includegraphics[width=11cm]{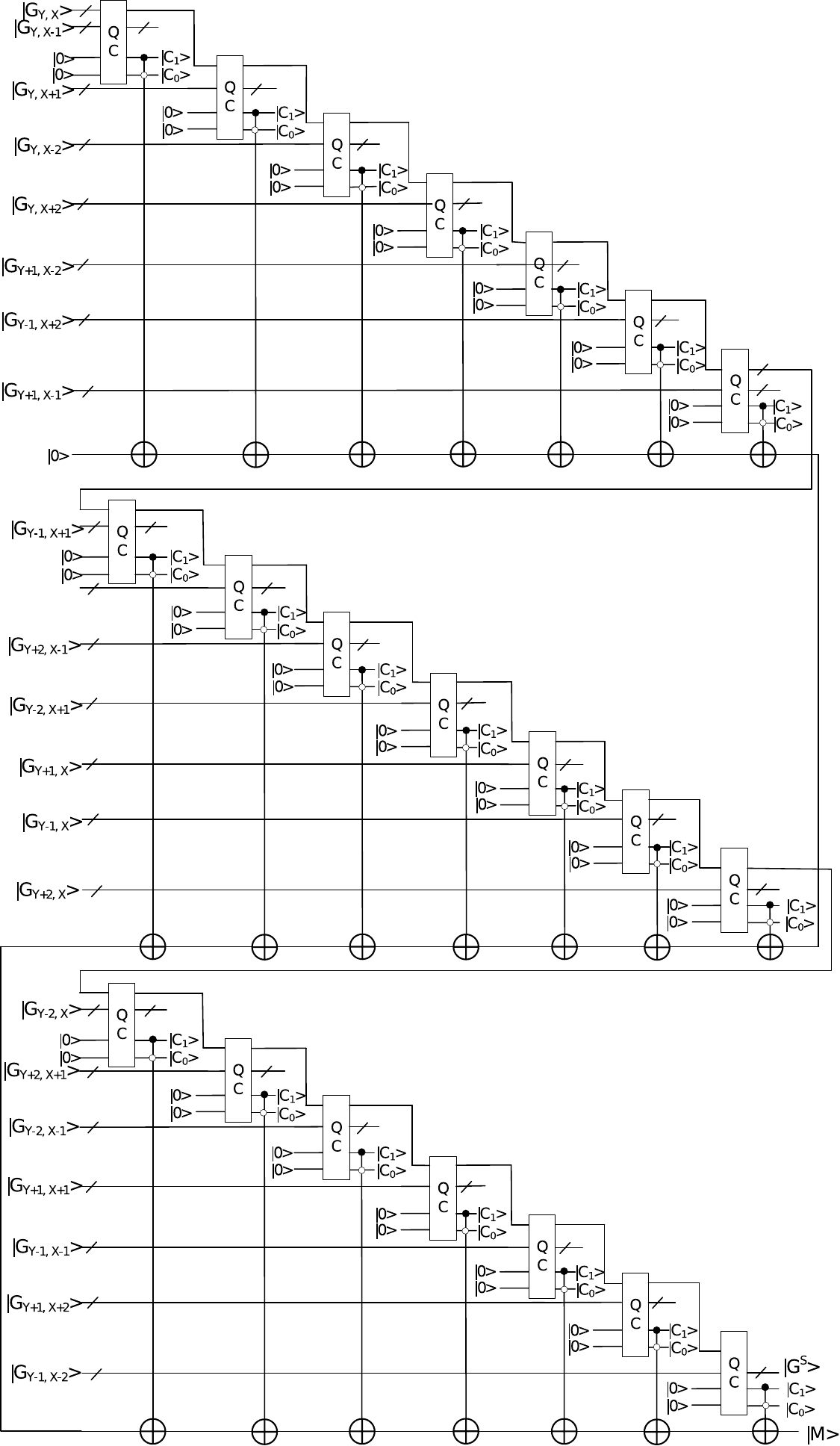}
    \caption{ Quantum circuit realization of non-maximum suppression}
    \label{Fig17}
\end{figure}

\begin{figure}
    \centering
    \includegraphics[width=11.5cm]{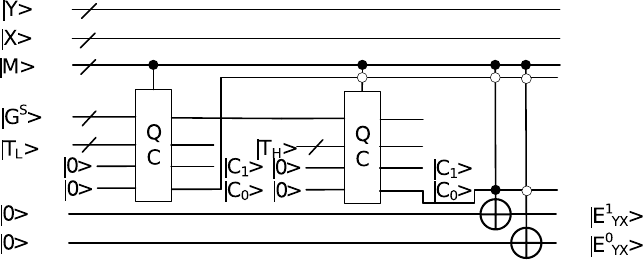}
    \caption{The quantum circuit realization of the double threshold detection}
    \label{Fig18}
\end{figure}

\noindent\textbf{Step 4 The non-maximum suppression.} Non-maximum suppression means setting the current pixel grayscale value to 0 if the gradient value is smaller than the two pixels' in its gradient direction, then the current pixel is a non-maximum pixel; If the gradient value of the current pixel is greater than or equal to the gradient value of the two pixel in its gradient direction, the current pixel is determined as a maximum point, and it is to be retained. In this way, the points with the maximum local gradient values can be retained, which can eliminate edge false detections. In this paper, we use the Sobel operator to calculate the gradient values for all eight directions (${0^ \circ }$, ${22.5^ \circ }$, ${45^ \circ }$, ${67.5^ \circ }$, ${90^ \circ }$, ${112.5^ \circ }$, ${135^ \circ }$, ${157.5^ \circ }$). Quantum comparators are used to find the maximun local gradient value pixels $\lvert {{G^s}} \rangle $ of the gradient image $\lvert {{G}} \rangle $ obtained with the Sobel operator in eight directions. Each pixel's information is obtained from the $5\times5$ neighborhood window using the previously prepared NEQR image set.   Quantum gradient image $\lvert {{G^S}} \rangle $ after non-maximum suppression can be written as:
$$\lvert {{G^S}} \rangle  = \frac{1}{{{2^n}}}\sum\limits_{Y = 0}^{{2^n} - 1} {\sum\limits_{X = 0}^{{2^n} - 1} {\lvert M \rangle \lvert {{G_{Y\!X}}} \rangle \lvert Y \rangle \lvert X \rangle } }\eqno{(13)} $$
where $\lvert M \rangle =1$ indicates that the current pixel is a maximum pixel point and $\lvert M \rangle =0$  indicates that the current pixel point is a non-maximum pixel point. Fig. \ref{Fig17} presents the quantum circuit design for non-maximum suppression.
\noindent\textbf{Step 5 Double threshold detection}. After non-maximum suppression, the remaining pixels can more accurately represent the edges in the image. But it will still be affected by some noise present. To address the problem, the double threshold  need to be used for detection. The high threshold $T_H$ and the low  threshold $T_L$  are selected to divide the edge points. Pixels with gradient values less than the low threshold are determined as non-edge points, pixels with gradient values between high threshold and low threshold are determined as weak edge points, and pixels with gradient values greater than the high threshold are determined as strong edge points. All pixels' gradient values  of the $5\times5$ neighborhood are  compared with the double threshold, where the relationship between the high and low threshold is $\lvert {{T_L}} \rangle  = \frac{1}{3}\lvert {{T_H}} \rangle $. The quantum image obtained after  double threshold detection can be  expressed as: 
$$\lvert E \rangle=\frac{1}{{{2^n}}}\sum\limits_{Y = 0}^{{2^n} - 1} {\sum\limits_{X = 0}^{{2^n} - 1} {\lvert {{E_{Y\!X}}} \rangle } } \lvert Y \rangle \lvert X \rangle \eqno{(14)}$$
where ${E_{Y\!X}}\!=\!{E^0{_{Y\!X}}}{E^1{_{Y\!X}}},{E^h{_{Y\!X}}} \in \left\{ {0,1} \right\}, h\! =\! 0,1$. The correspondence between the values of the ${E_{YX}}$ and the 3 kinds of edge points are: if ${E_{Y\!X}} \!= \!10$, then it is a strong edge point; if ${E_{Y\!X}}\! = \!01$, then it is a weak edge point; if ${E_{Y\!X}} \!=\! 00$, then it is not an edge point. The quantum circuit of double threshold detection  is shown in Fig. \ref{Fig18}. 

\begin{figure}
    \centering
    \includegraphics[width=11cm]{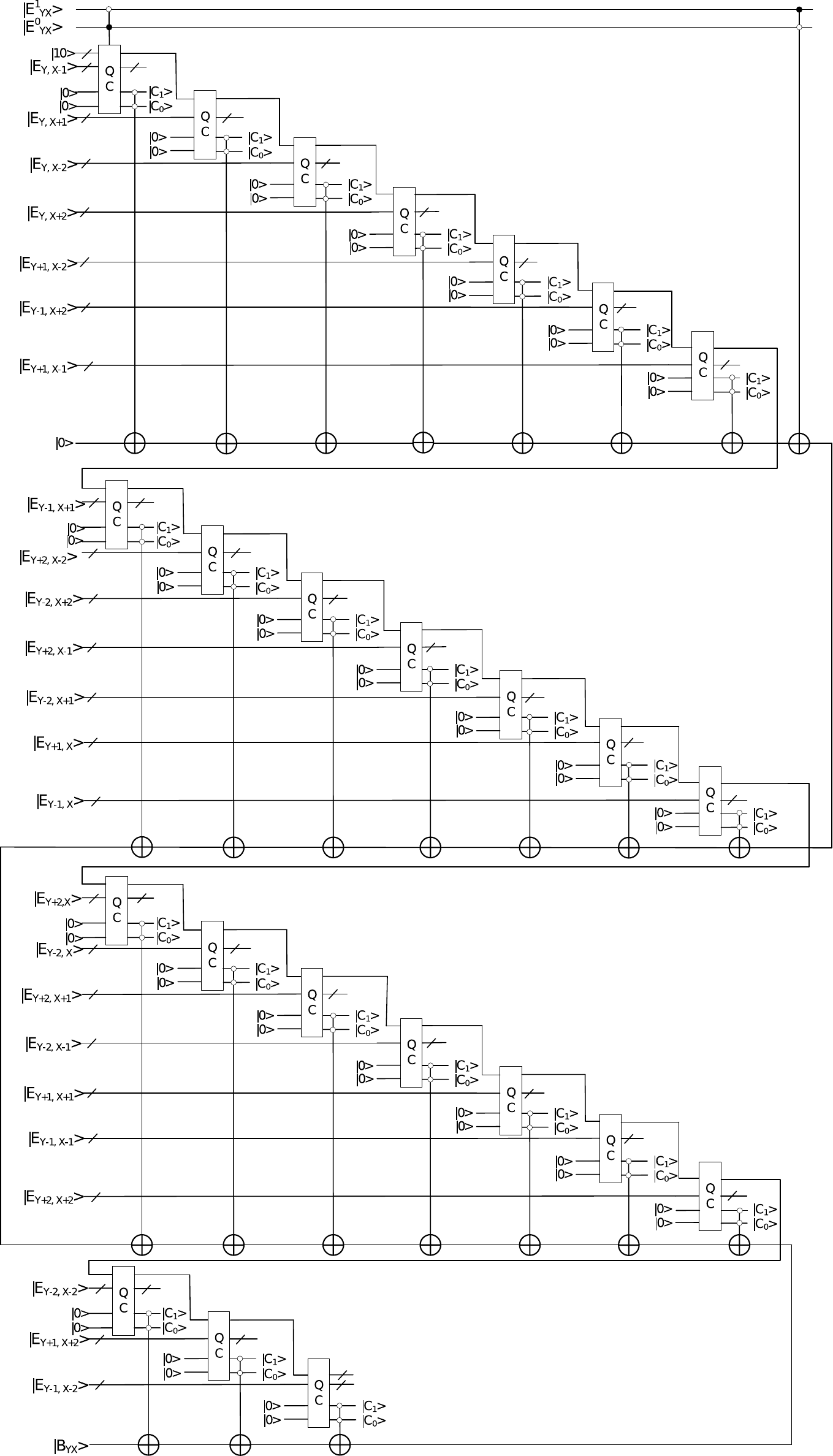}
    \caption{The quantum circuit implementation of edge tracking}
    \label{Fig19}
\end{figure}

\noindent\textbf{Step 6 Edge tracking}. After double threshold detection, the pixels  classified as strong edge points have been determined as edges because these edges are real edges in the image. The weak edge points may be real edges or caused by factors such as noise, which requires further processing through edge tracking. When a strong edge point exists in the 24 neighborhood of a pixel centered on the weak edge point, the weak edge point is determined as a true edge point left, otherwise the weak edge point is determined as a false edge point. Based on the division of strong and weak edge points in the fifth step, if ${E_{Y\!X}} = 01$, then the current pixel is a weak edge point. Under the auxiliary qubits control, the double threshold detection information of the $5\times5$ neighborhood pixels was obtained using the cyclic shift operation, and the double threshold detection results of each pixel of the 24 neighborhood were compared using the quantum comparator. The double threshold detection results for each pixel in the neighborhood were then compared with the quantum sequence $\lvert {00} \rangle$ for the presence of strong edge points in the 24 neighborhood using the auxiliary qubits $\lvert {B_{Y\!X}}\rangle$. If ${B_{Y\!X}=1}$, this indicates the presence of strong edge points, which otherwise do not exist. The final quantum state of the quantum edge image after the edge tracking operation are represented as follows:
$$\lvert B  \rangle =\frac{1}{{{2^n}}}\sum\limits_{Y = 0}^{{2^n} - 1} {\sum\limits_{X = 0}^{{2^n} - 1} {\lvert {{B_{Y\!X}}} \rangle } } \lvert Y \rangle \lvert X \rangle \eqno{(15)}$$
where ${B_{Y\!X}=1}$ in case of edge point and ${B_{Y\!X}=0}$ in case of non-edge point. The quantum circuit implementation of edge tracking is shown in Fig. \ref{Fig19}.

\section{Circuit complexity and experiment analysis}\label{Section 4}
In this section, we first discuss the circuit complexity based on edge detection of the eight-direction Sobel operator and compare the complexity of our algorithm with some existing edge detection algorithms. Then, simulation experiments are presented to show the effect of edge detection in quantum images.
\subsection{Circuit complexity analysis}
The NOT gate and CNOT gate are commonly used in quantum computing. This paper considers its circuit complexity as  1. Therefore, we can compute the complexity of the quantum circuit with the number of basic logic gates.  In Ref. \cite{Nielsem2000},  Nielsem et al. point out that the Toffoli gates of 3 qubits can be decomposed into five two-qubit gates, so the complexity of the Toffoli gate is 5. The CNOT gate ${{\rm{C}}_{n - 1}}(x)$ (The number of control qubits is $n-1$) can be decomposed into the quantum circuit with $2(n-1)$ Toffoli gates and 1 CNOT gate \cite{Nielsem2000}. Thus the ${{\rm{C}}_{n - 1}}(x)$ gate circuit complexity is $10n-9$.

Taking an image of size $2^n\times2^n$ as an example, we discuss the complexity of the circuit in six steps. They are quantum image preparation, Quantum image set cyclic shift, gradient  calculation based on eight-direction Sobel operator, non-maximum suppression, double threshold detection and edge tracking.

In step 1, NEQR  image is prepared. Digital image is prepared as NEQR quantum image. The computational complexity of this step is O$(qn2^{2n})$ \cite{Zhang2013}.

In step 2, the quantum image is cycle shifted. This step requires Copy operations \cite{Fan2019} and CT operations \cite{Le2010,Le2011S,Wang2014}. The complexity  is O$(n^2)$ \cite{Wang2014}.

In step 3, the gradients  are calculated. This step is to calculate the  gradient  of each pixel. The quantum adder, quantum double operation, absolute value operation, quantum comparator and swap operation are needed. The complexity of each q-qubit quantum adder operation and the quantum  double operation are O$(q)$ \cite{Li2018B}. The circuit complexity of the absolute value operation is O$(q^2)$ \cite{Fan2019,Thapliyal2009}. The quantum comparator has a complexity of O$(n)$ \cite{Oliveira2007}. The complexity of the Swap operation is O$(n)$ \cite{Li2018B}. Therefore, the circuit complexity of this step is O$(n+q^2)$.

In step 4, non-maximum pixels are suppressed. The 25 additional images and 5×5 neighborhood window pixels are replicated with the Copy operation and then cycle shifted with the CT operation. In addition, this step requires  quantum comparators and  Toffoli gates to find the maximum gradient   value pixel. Therefore, the circuit complexity in this step is O$(n^2)$.

In step 5, double threshold is used to compare edge pixels. The quantum comparator, Toffoli gate and CNOT gate are used. Therefore, the circuit complexity at this step is O$(n)$.

In step 6, edge pixels are edge-tracked. This step requires CT operation,  quantum comparators  and some Toffoli gates. Therefore, the circuit complexity at this phase is O$(n^2)$.

According to the complexity analysis of the above 6 steps, we can know that the computational
complexity of circuit realization of  QSED   for a $2^n \times 2^n$  classical image is 
\begin{align*}
    &{\rm{O}}[qn2^{2n}+n^2+(n+q^2)+n^2+n+n^2]\\
    &={\rm{O}}(qn2^{2n}+n^2+q^2)
\end{align*}
The QIP algorithm is for quantum images rather than classical images, but it is currently impossible to directly obtain quantum images, so we  need to convert the classical images into quantum images firstly. For the completeness of the paper, we also analyze the complexity of the quantum image preparation process. But typically, the  quantum image preparation and measurement processes are not considered part of quantum image processing \cite{Fan2019,Chetia2021}. Therefore, for $2^n\times2^n$ images, the  complexity of our algorithm  is O$(n^2 +q^2)$.  On classical computers, for images of size $2^n\times2^n$, edge detection need to be processed individually for each pixel. So, the complexity of the classical edge detection algorithm is no less than  O$(2^{2n})$ \cite{Fan2019}. Thus, our scheme achieves an exponential acceleration relative to the classical edge detection algorithm, so the real-time problem in classical image edge detection can be solved well. In Table \ref{tab2}, the computational complexity of our algorithm is compared with some other edge detection schemes, and the complexity of our algorithm is greatly improved.
\begin{table}[h]
\begin{center}

\caption{Comparison of the complexity of the Sobel edge detection algorithm}\label{tab2}%
\resizebox{.80\columnwidth}{!}{
\begin{tabular}{llll}
\toprule
Algorithm & Encoding model  & Complexity & Directions\\
\midrule
Sobel \cite{Zheng2013,Nielsem2000,Rosenfeld1976}        & -      & O$(2^{2n})$  & 2/4/8  \\

Fan \cite{Fan2019}        & NEQR     & O$(n^2+2^{q+4})$  &2 \\

R.Chetia \cite{Chetia2021}   & NEQR    & O$(n^2+ q^3)$   &4\\
Our scheme    & NEQR    & O$(n^2+ q^2)$   &8\\
\botrule
\end{tabular}
}

\end{center}
\end{table}

\subsection{Experiment analysis}
Due to the limitations of the current technology, there are no suitable quantum computers for our use. To test our proposed scheme, all experiments were simulated on a classical computer with MATLAB 2014. The unit vector and unitary matrices in MATLAB can replace that of the quantum states and quantum gates, respectively. Therefore, although the simulation on a classical computer can not truly realize the quantum model simulation, it can simulate the execution steps of quantum computation, which can theoretically verify the effectiveness of the quantum algorithm. 

Five common test images were selected randomly, such as Lena, cameraman,  Livingroom, House and Pirate. The size of the images is 512×512. We compare the quantum two-direction and   four-direction Sobel operator edge detection algorithm   with our proposed eight-direction Sobel operator edge detection algorithm.

\begin{figure}
    \centering
  \subfigure[]{ \includegraphics[width=2.5cm]{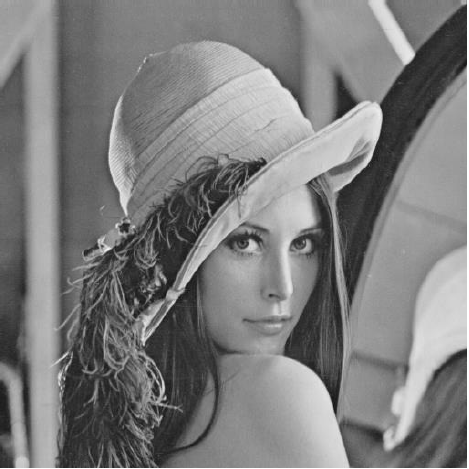}}
   \subfigure[]{\includegraphics[width=2.5cm]{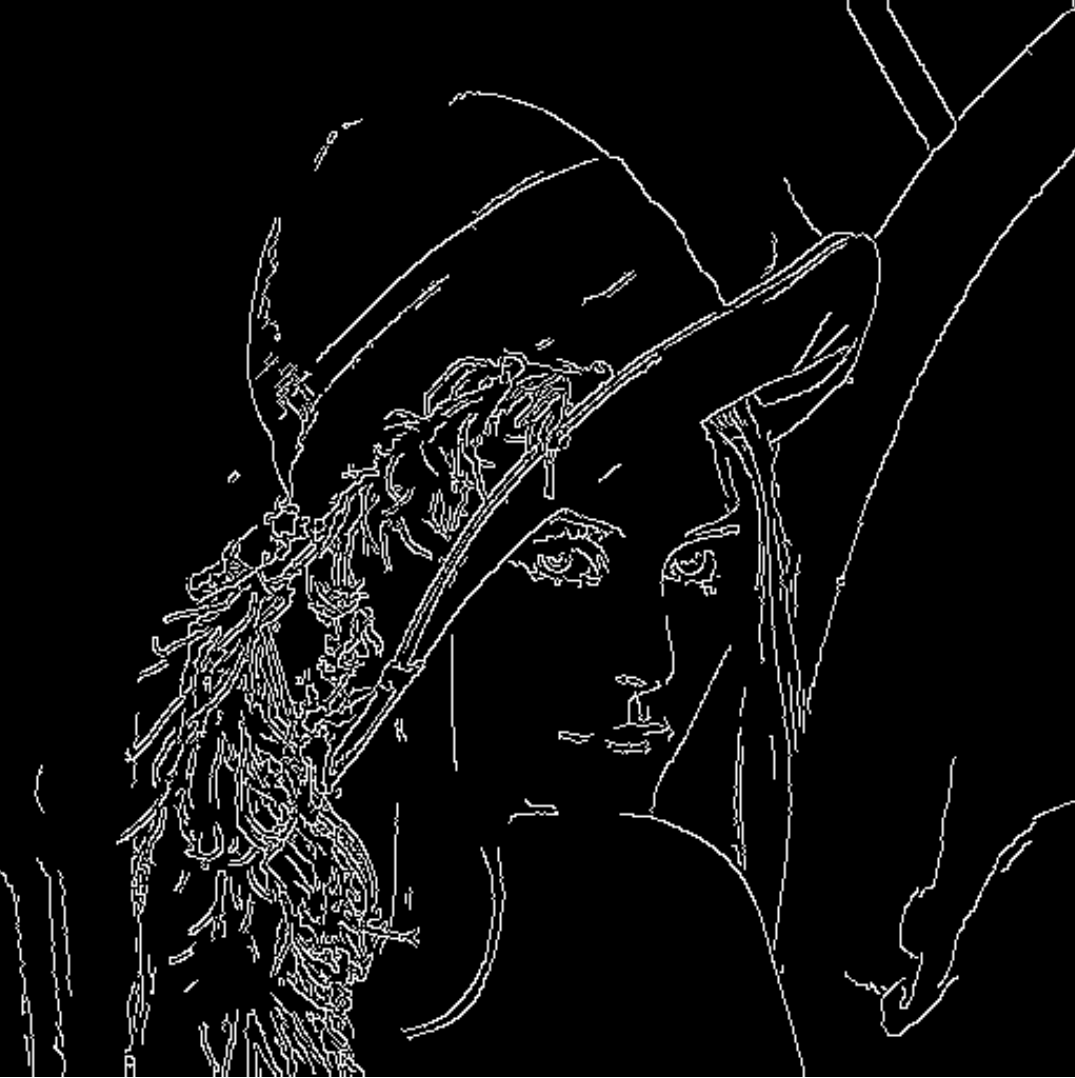}}
   \subfigure[]{\includegraphics[width=2.5cm]{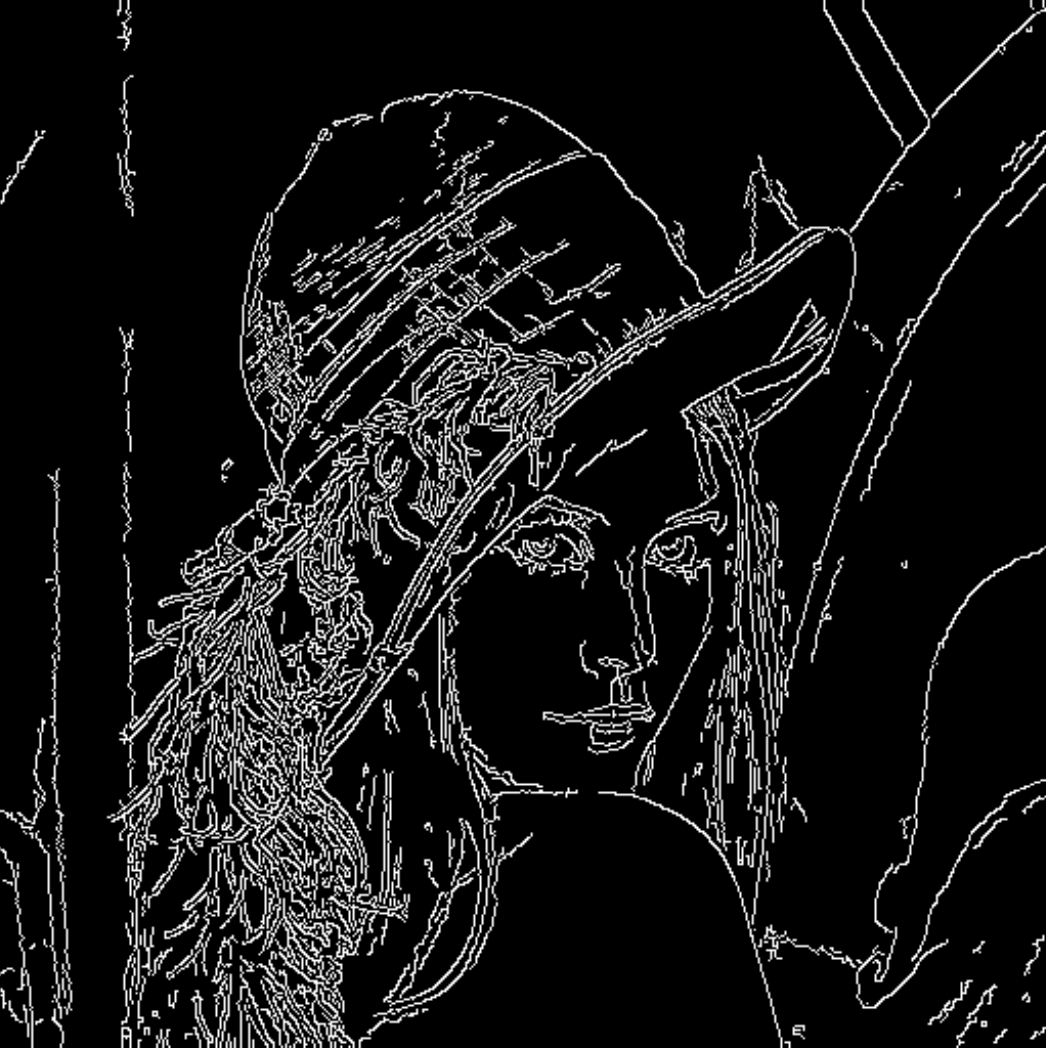}}
   \subfigure[]{\includegraphics[width=2.5cm]{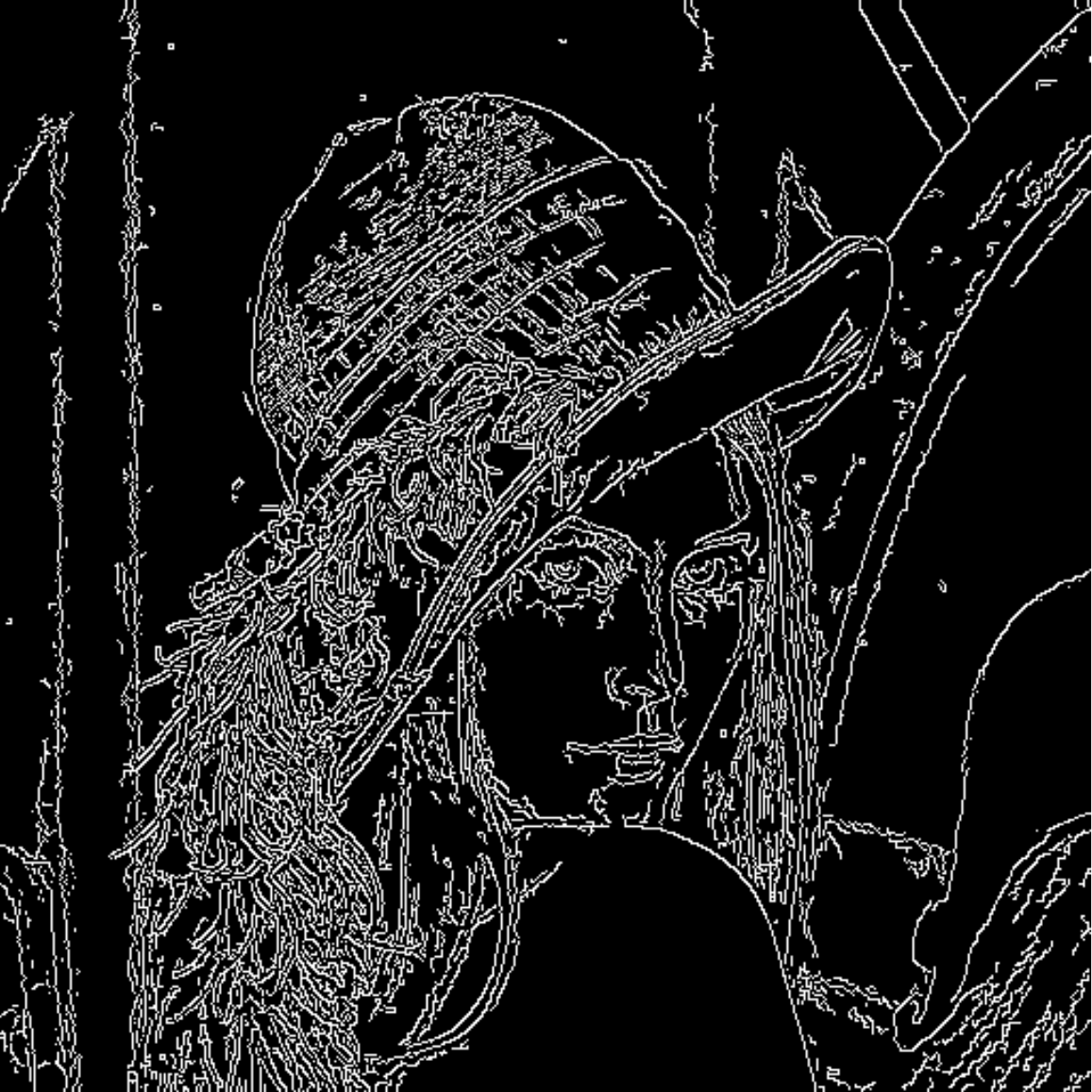}}\\
   \setcounter{subfigure}{0}\subfigure[]{\includegraphics[width=2.5cm]{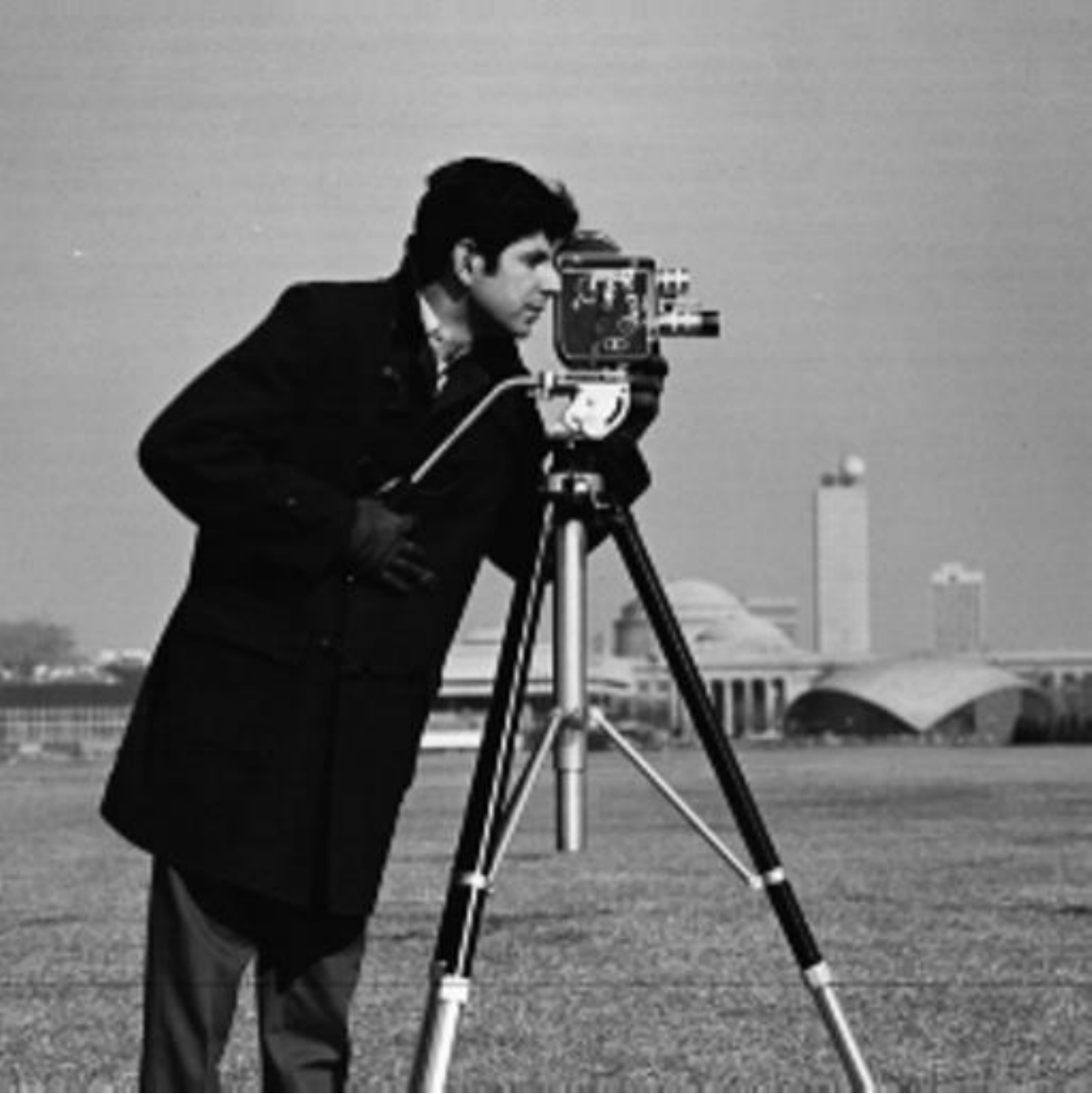}}
   \subfigure[]{\includegraphics[width=2.5cm]{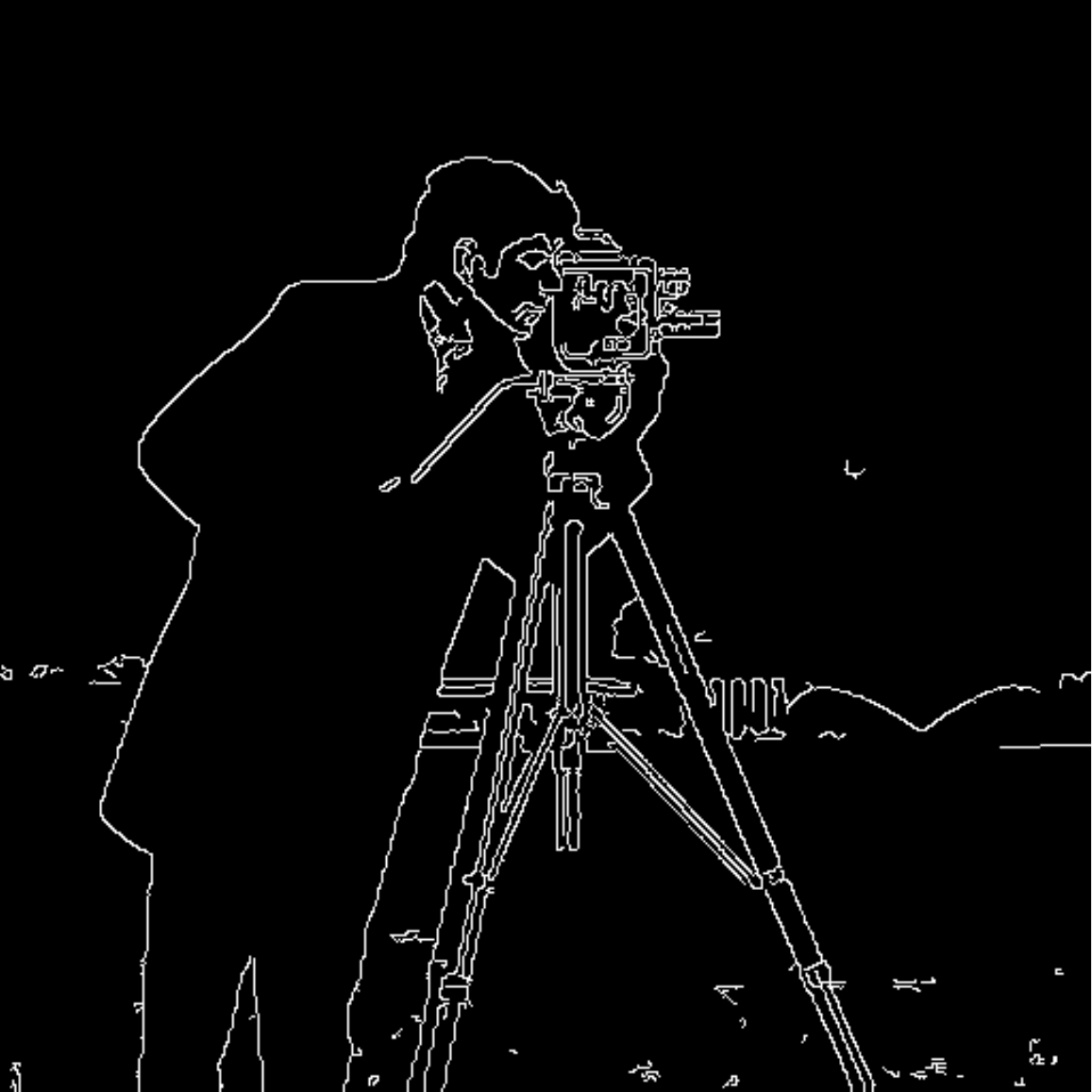}}
   \subfigure[]{\includegraphics[width=2.5cm]{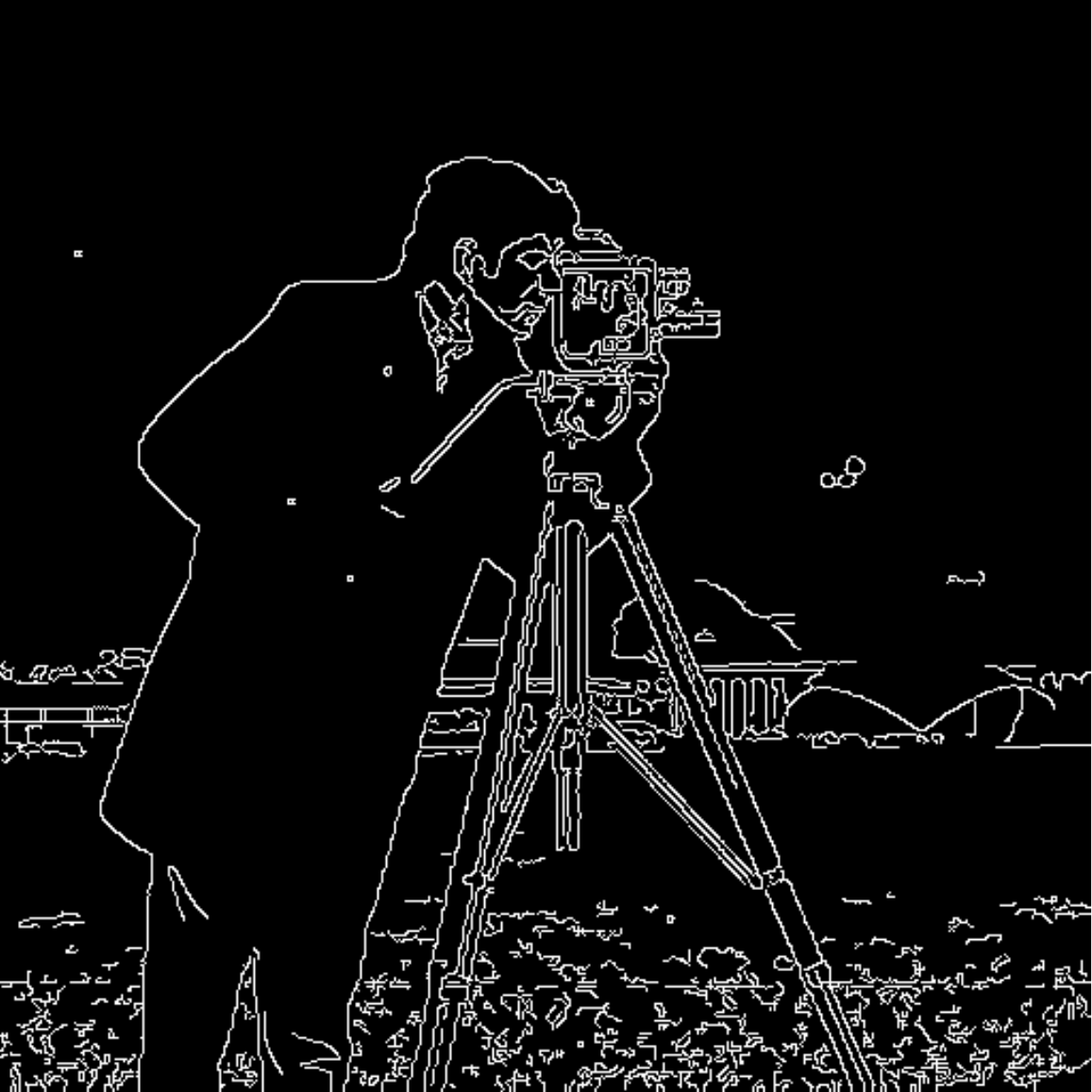}}
   \subfigure[]{\includegraphics[width=2.5cm]{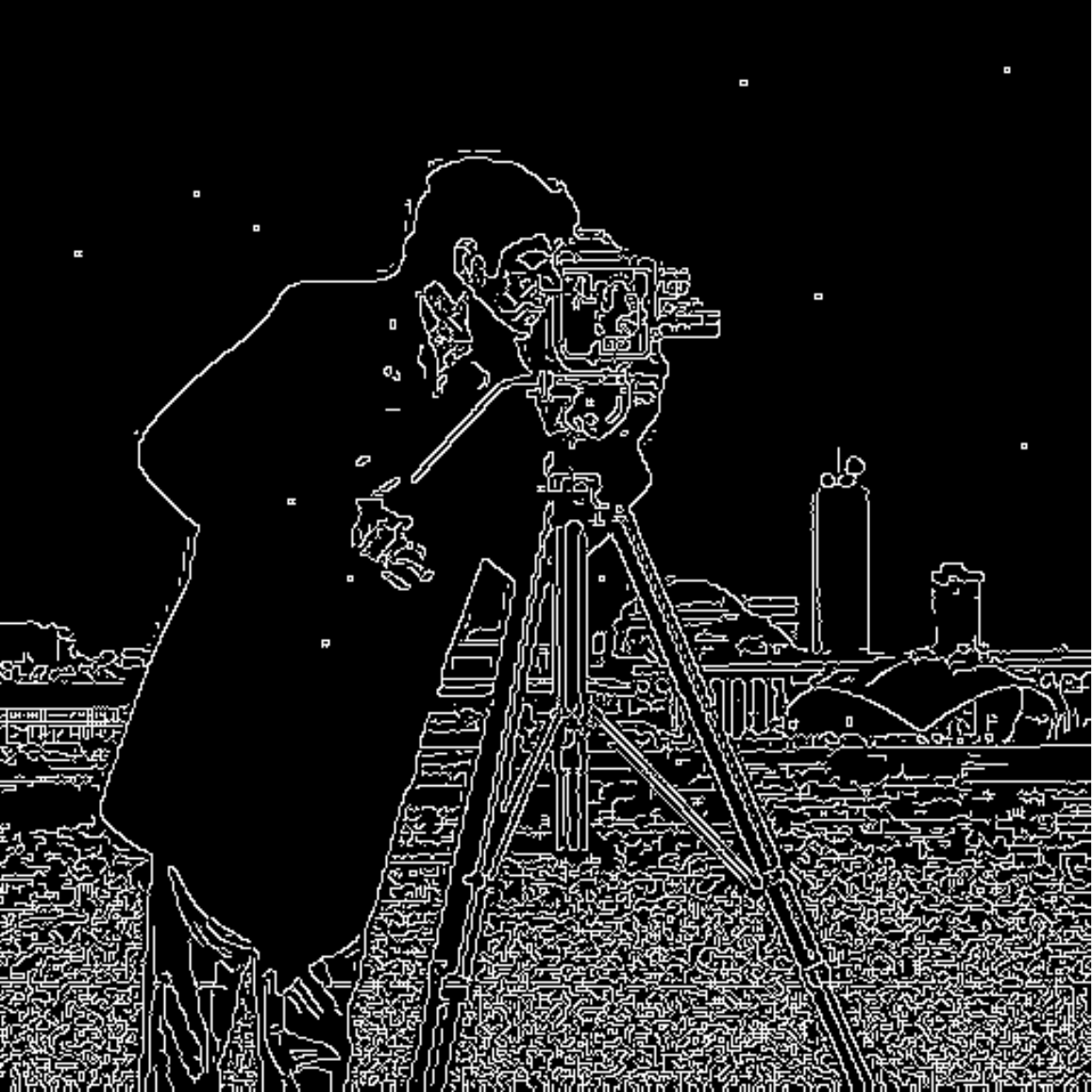}}\\
   \setcounter{subfigure}{0}\subfigure[]{\includegraphics[width=2.5cm]{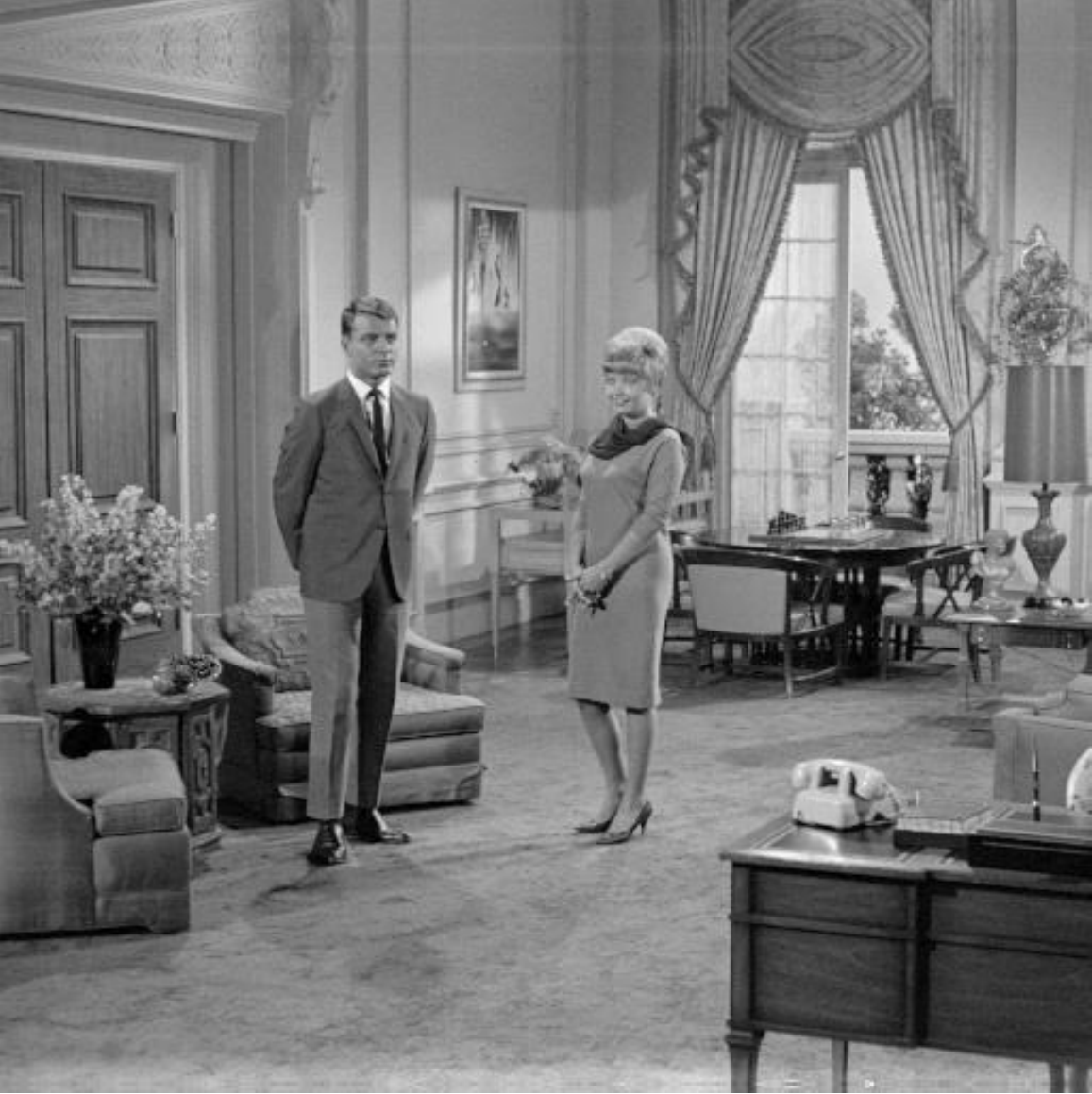}}
   \subfigure[]{\includegraphics[width=2.5cm]{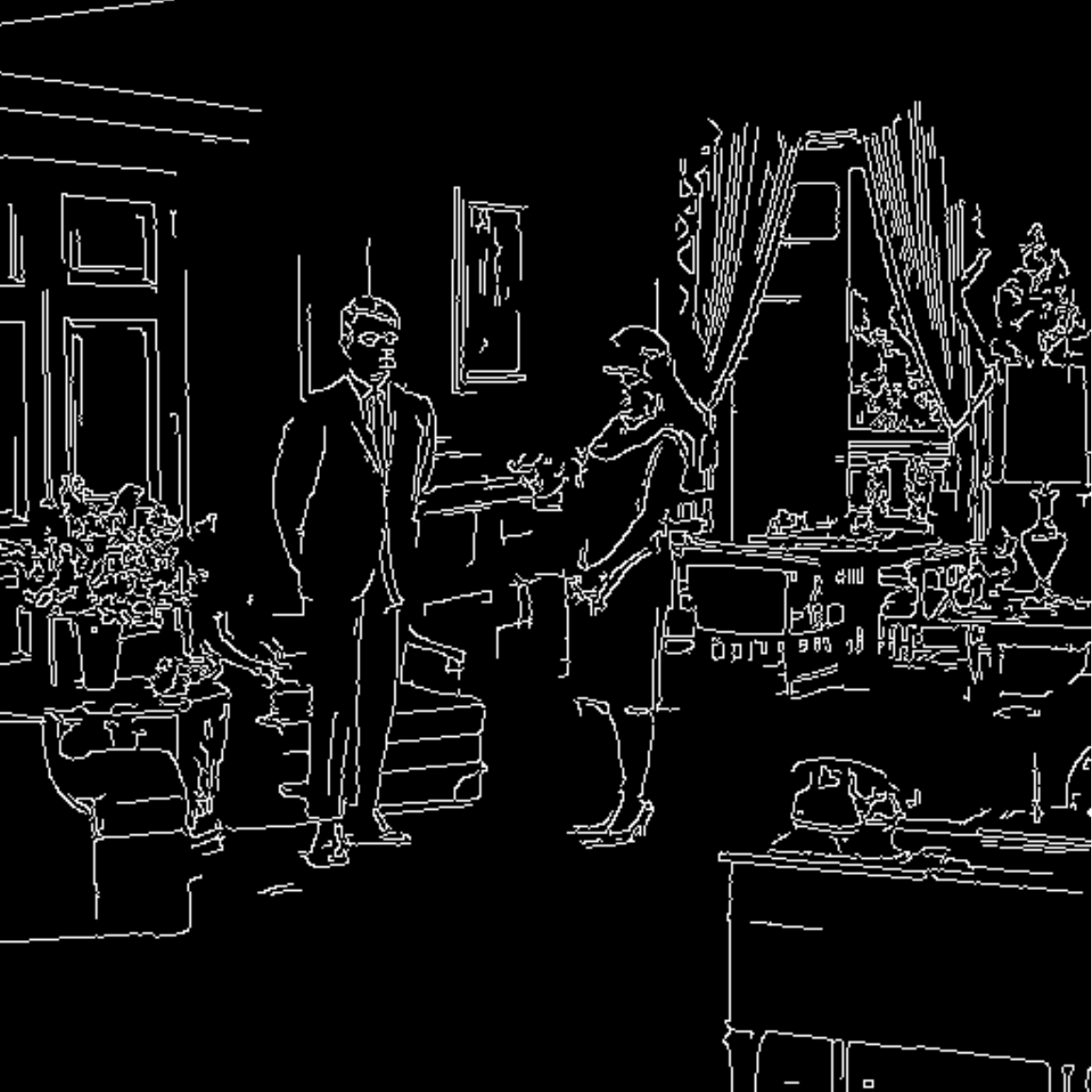}}
   \subfigure[]{\includegraphics[width=2.5cm]{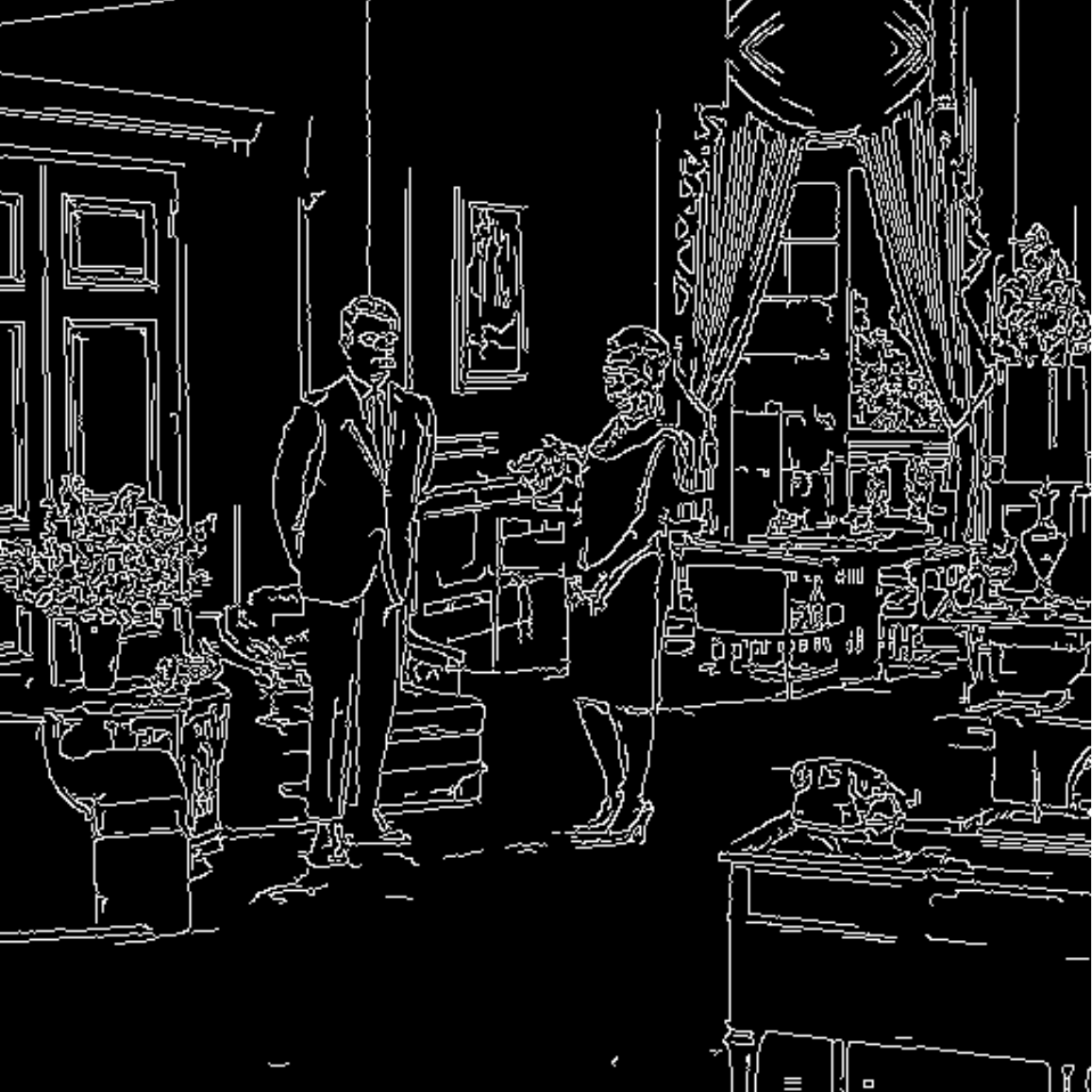}}
   \subfigure[]{\includegraphics[width=2.5cm]{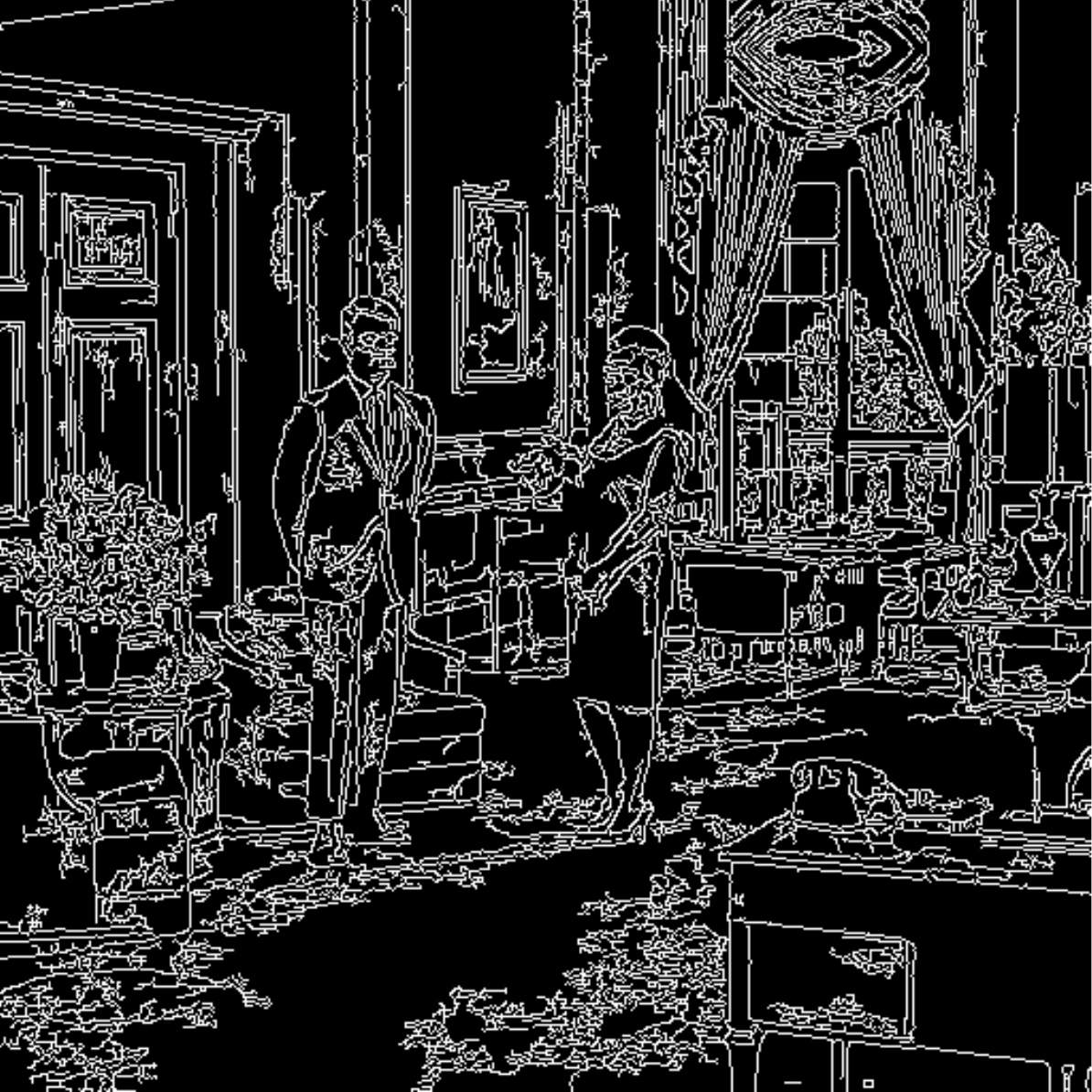}}\\
  \setcounter{subfigure}{0}\subfigure[]{\includegraphics[width=2.5cm]{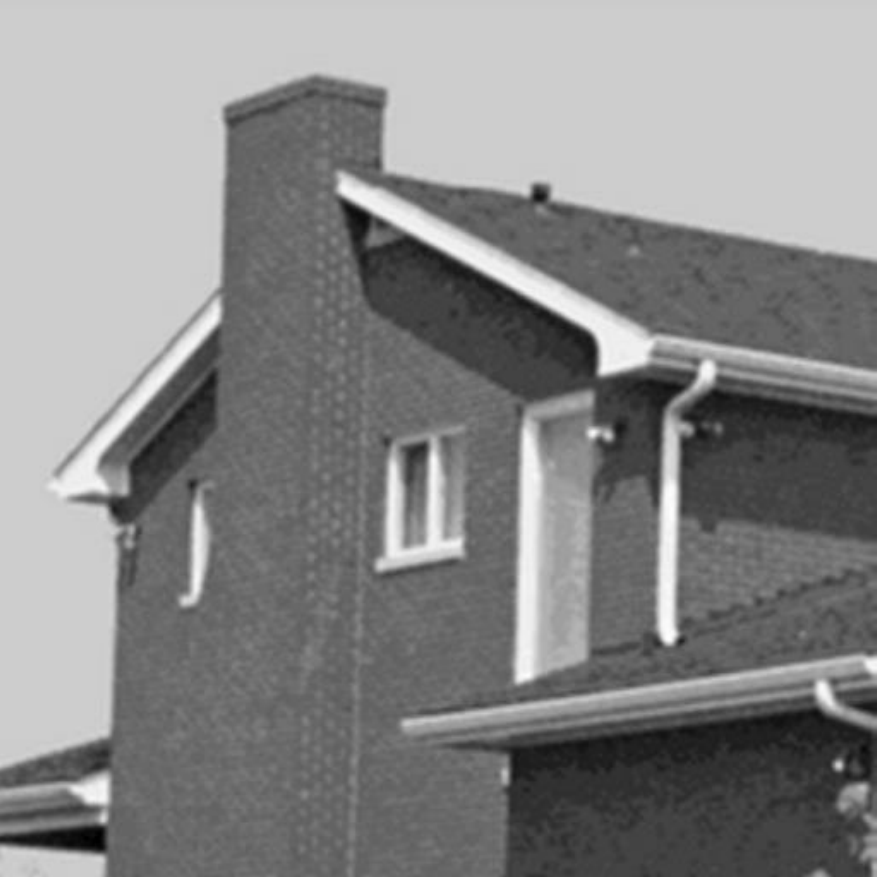}}
   \subfigure[]{\includegraphics[width=2.5cm]{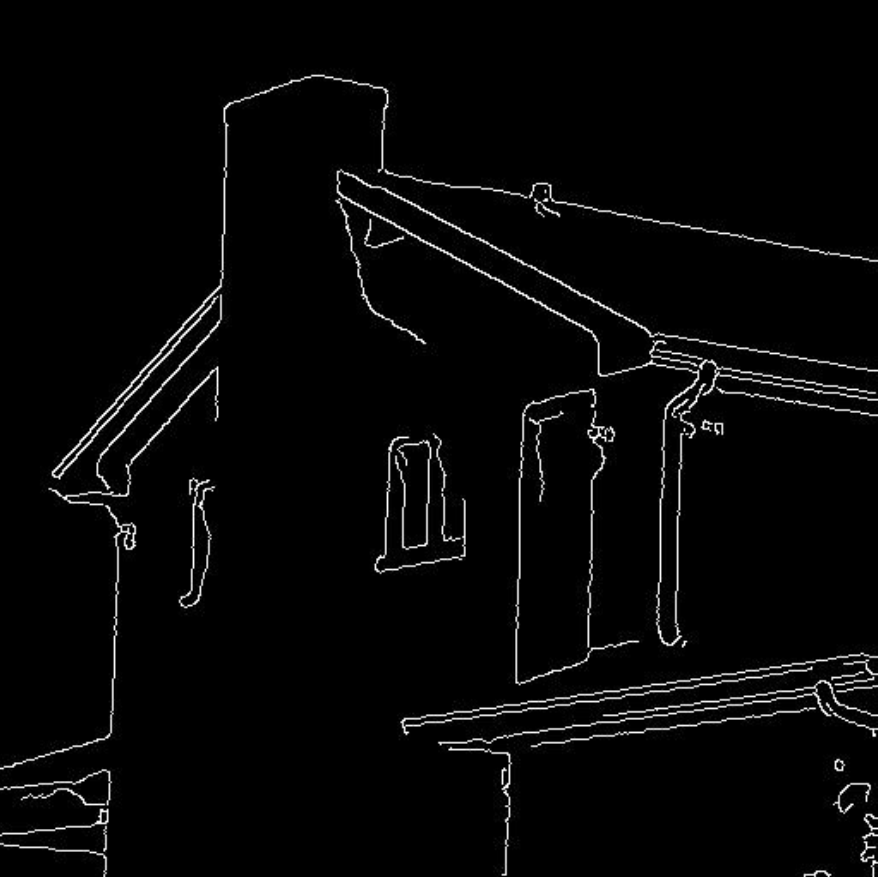}}
   \subfigure[]{\includegraphics[width=2.5cm]{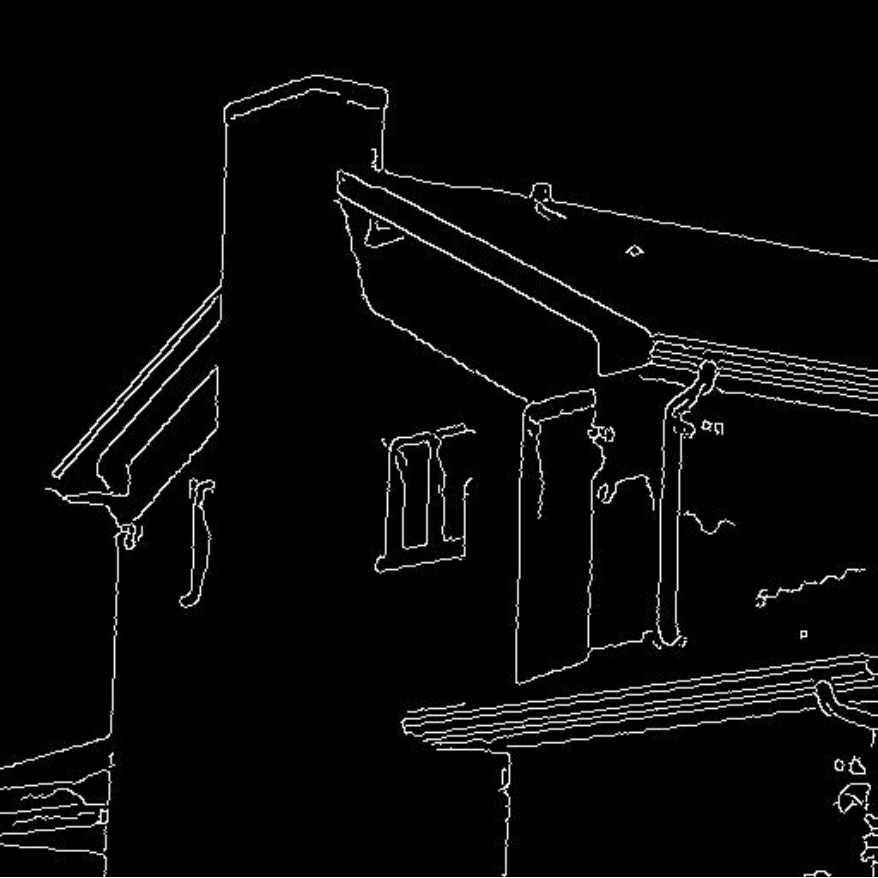}}
   \subfigure[]{\includegraphics[width=2.5cm]{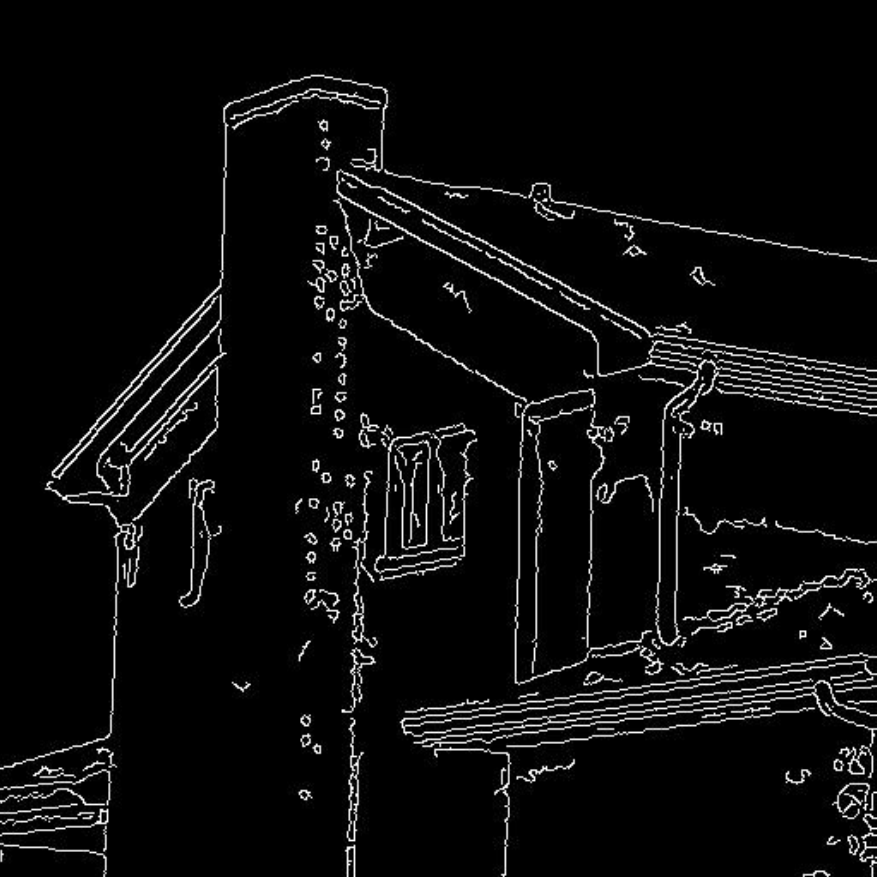}}\\
    \setcounter{subfigure}{0}\subfigure[]{\includegraphics[width=2.5cm]{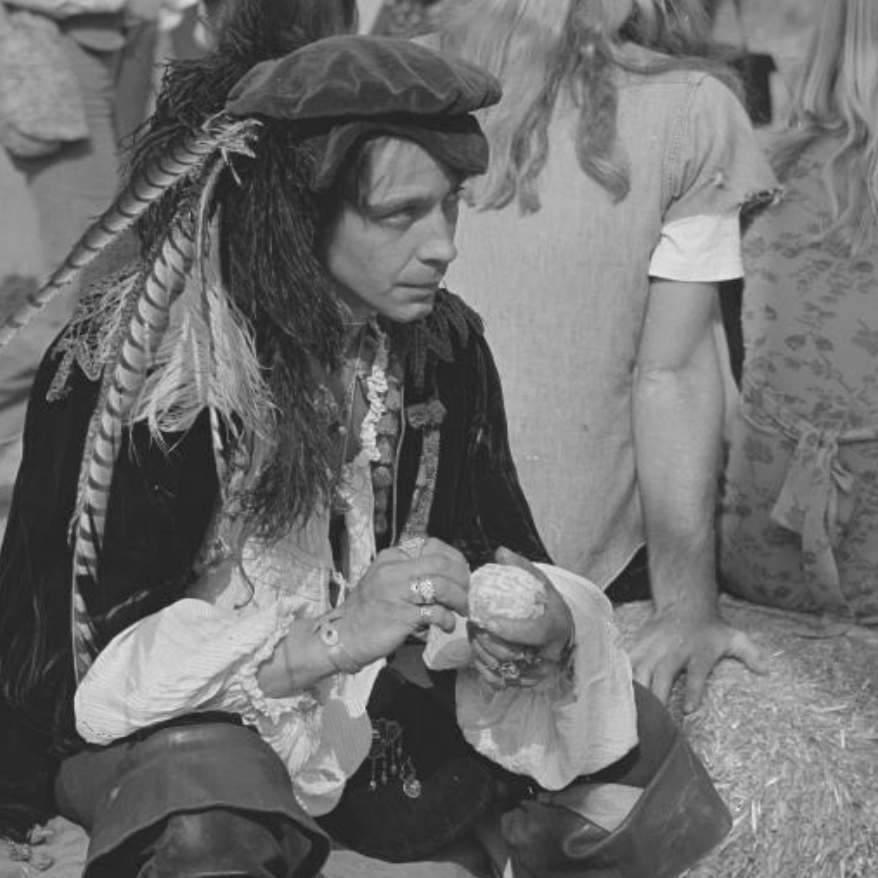}}
   \subfigure[]{\includegraphics[width=2.5cm]{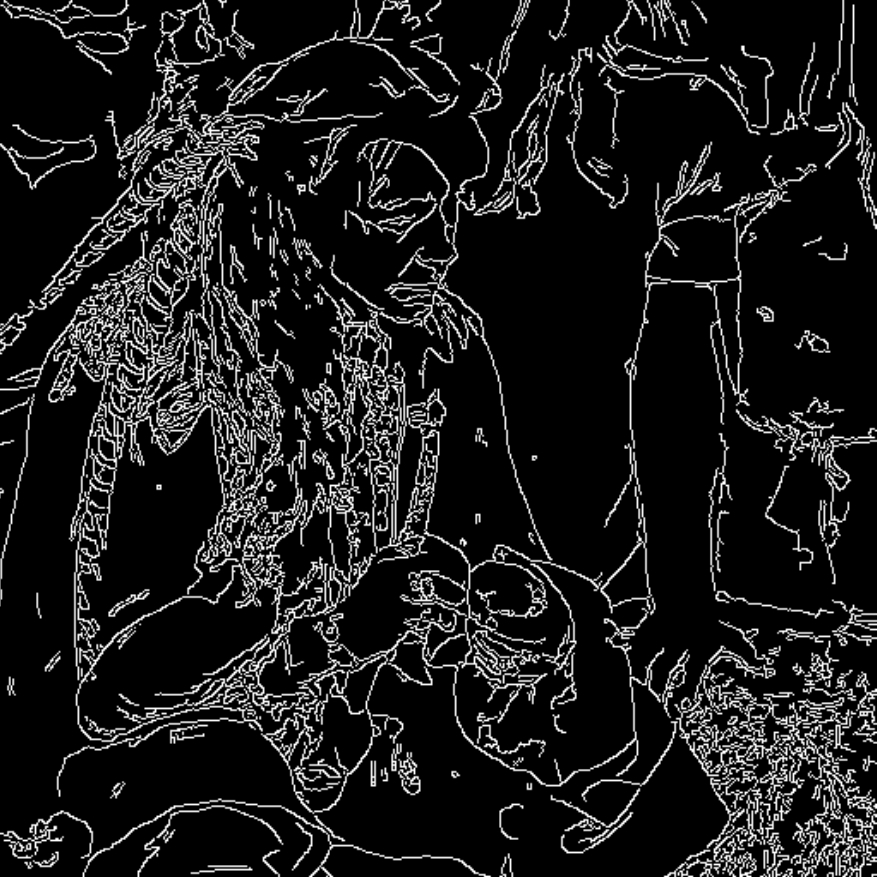}}
   \subfigure[]{\includegraphics[width=2.5cm]{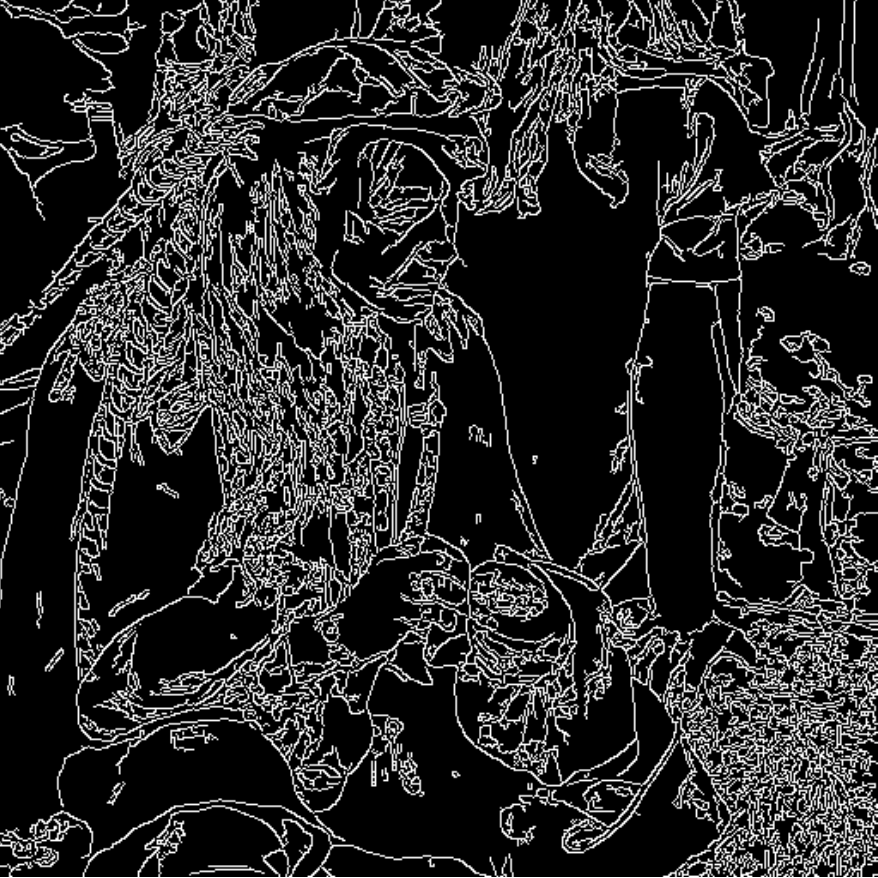}}
   \subfigure[]{\includegraphics[width=2.5cm]{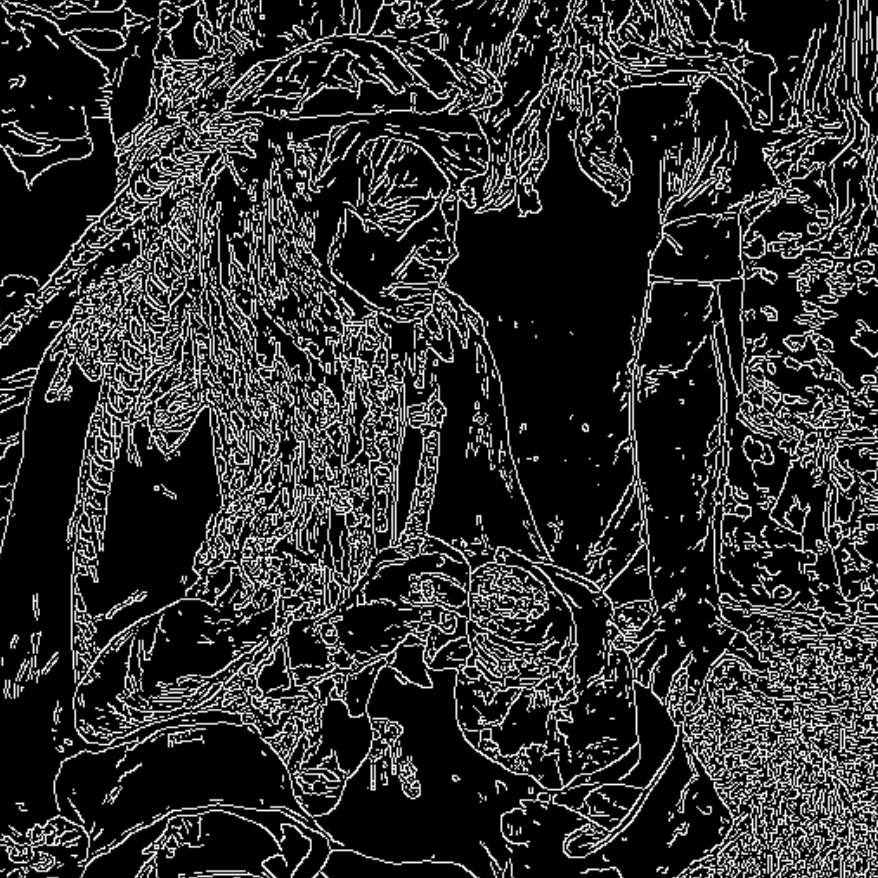}}\\
   
    \caption {a Five common and original test images. b The result images of  the two-direction Sobel operator edge detection algorithm. c The result images of  the four-direction Sobel operator edge detection algorithm. d The result images of our proposed  algorithm.}
    \label{fig20}
\end{figure}

As can be seen from Fig. \ref{fig20},  our algorithm detects more edge information, especially in the more detailed parts, such as Leana's hat, the photographer's grass,  the livingroom's curtains, the house's wall and the pirate' hair accessories. This is because we employ the Sobel mask of 5×5 to detect image edges from eight directions and further process edge information using non-maximum suppression double threshold  values detection and edge tracking, from which we obtain a clearer edge profile and more edge information. 

In addition, we also use the mean square error (MSE) to judge the quality of the resulting image, which is one of the most commonly used methods for judging image quality. In this paper, the fewer false edges in the detected image, the smaller MSE value. For two gray-scale images Q and R with size $2^n\times2^n$, MSE is defined as 
\[MSE=\frac{1}{2^{2n}} \sum_{Y=0}^{2^{n}-1} \sum_{X=0}^{2^{n}-1}[Q(Y, X)-R(Y, X)]^{2}\eqno{(16)}\]
where Y and X represent the position information of the images.

From Tab. \ref{tab3}, it can be seen that the   MSE values of all  images detected by our algorithm is less than that of images detected by the other two algorithms, which is because our algorithm detects fewer false edges. To sum up, our algorithm can not only detect more edge pixels, but also detect fewer false edges, which is meaningful.

\begin{table}[h]
\begin{center}

\caption{Comparison of the MSE values of the different QSED algorithms}\label{tab3}%
\resizebox{.80\columnwidth}{!}{
\begin{tabular}{llll}
\toprule
\multicolumn{1}{c}{\multirow{2}{*}{Input image}} & \multicolumn{3}{c}{MSE}                                    \\ \cmidrule(l){2-4} 
\multicolumn{1}{c}{}                             & Two-direction QSED & Four-direction QSED & Our algorithm \\ 
\midrule

Lena      & 159.16      & 153.19  & 147.27  \\

Cameraman        &186.05     & 183.06  &181.58  \\

Livingroom   & 169.32    & 167.88   &164.80\\
House    & 217.95    & 217.26  &216.01\\
Pirate    &159.68     & 158.39   &154.49\\

\botrule
\end{tabular}
}

\end{center}
\end{table}

\section{Conclusion}\label{Section 5}
In this paper, based on the eight-direction Sobel operator, a novel quantum image edge detection algorithm is proposed, which can simultaneously calculates eight directions’ gradient values of all pixel in a quantum image. In addition, it combines non-maximum suppression, double threshold detection and edge tracking, which  can detect more accurate edge information. The concrete quantum circuits realization  are reported that our  algorithm can detect edges in the  complexity of O(${n^2} + {q^2}$) for a NEQR image with a size of ${2^n} \times {2^n}$. Compared with the classical and some existing QSED algorithms, our algorithm can achieve a significant improvement in the case of edge information and circuit 
complexity. 

At present, the number of qubits of quantum computers available  is relatively small, which cannot meet the requirements of a certain scale quantum image processing, therefore   we  performed a  experimental  simulation on a classic computer in this paper.  In addition, we performed experimental simulations in an ideal scenario, and do not consider the effects of noise. How to introduce noise into our scenario and design an anti-noise QSED algorithm is our future research work.

\bmhead{Acknowledgments}
This work is supported by the National Natural Science Foundation of China (62071240, 61802175), the Natural Science Foundation of Jiangsu Province (BK20171458), and the Priority Academic Program Development of Jiangsu Higher Education Institutions (PAPD).

All data generated or analysed during this study are included in this published article [and its supplementary information files].




\end{document}